%% file: 6123.tex
\documentclass[a4paper,10pt,twocolumn,letterpaper]{article}

\usepackage{cvpr}              

\input{preamble}

\definecolor{cvprblue}{rgb}{0.21,0.49,0.74}
\usepackage[pagebackref,breaklinks,colorlinks,citecolor=cvprblue]{hyperref}
\usepackage{multicol}
\usepackage{multirow}
\def\paperID{6123} 
\def\confName{CVPR}
\def\confYear{2024}

\title{Content-Style Decoupling for Unsupervised Makeup Transfer \\ without Generating Pseudo Ground Truth}

\author{Zhaoyang Sun\textsuperscript{\rm 1} \quad
Shengwu Xiong\textsuperscript{\rm 1,2,3,4} \quad
Yaxiong Chen\textsuperscript{\rm 1} \quad
Yi Rong\textsuperscript{\rm 1 \dag} \\
\textsuperscript{\rm 1}Wuhan University of Technology \quad
\textsuperscript{\rm 2}Wuhan Huaxia Institute of Technology \\
\textsuperscript{\rm 3}Shanghai AI Laboratory \quad
\textsuperscript{\rm 4}Qiongtai Normal University \\
{\tt\small zhaoyangsun0304@outlook.com, \{xiongsw, chenyaxiong, yrong\}@whut.edu.cn}
}

\begin{document}
\maketitle
{\let\thefootnote\relax\footnote{{\textsuperscript{\dag}Corresponding author.}}}

\input{sec/0_abstract}
\input{sec/1_introduction}
\input{sec/2_related_works}

\input{sec/3_methodology}

\input{sec/4_experiment}

\input{sec/5_conclusion}
\clearpage
{
    \small
    \bibliographystyle{ieeenat_fullname}
    \bibliography{main}
}
\input{sec/X_suppl.tex}


\end{document}

%% file: preamble.tex
\usepackage[dvipsnames]{xcolor}


%% file: sec/0_abstract.tex
\begin{abstract}
The absence of real targets to guide the model training is one of the main problems with the makeup transfer task. Most existing methods tackle this problem by synthesizing pseudo ground truths (PGTs). However, the generated PGTs are often sub-optimal and their imprecision will eventually lead to performance degradation. To alleviate this issue, in this paper, we propose a novel Content-Style Decoupled Makeup Transfer (CSD-MT) method, which works in a purely unsupervised manner and thus eliminates the negative effects of generating PGTs. Specifically, based on the frequency characteristics analysis, we assume that the low-frequency (LF) component of a face image is more associated with its makeup style information, while the high-frequency (HF) component is more related to its content details. This assumption allows CSD-MT to decouple the content and makeup style information in each face image through the frequency decomposition. After that, CSD-MT realizes makeup transfer by maximizing the consistency of these two types of information between the transferred result and input images, respectively. Two newly designed loss functions are also introduced to further improve the transfer performance. Extensive quantitative and qualitative analyses show the effectiveness of our CSD-MT method. Our code is available at https://github.com/Snowfallingplum/CSD-MT.
\end{abstract} 

%% file: sec/1_introduction.tex
\section{Introduction}
\label{sec:intro}
Given a pair of source and reference face images, the main goal of makeup transfer is to generate an image that simultaneously satisfies the following conditions:
(1) Containing the makeup styles transferred from the reference image, such as lipstick, eye shadow and powder blush.
(2) Preserving the content details of the source image, including identity, facial structure and background.
This technique has been widely studied and is attracting increasing attentions from the computer vision and artificial intelligence communities, due to its great economic potential in the fields of e-commerce, entertainment and beauty industries. However, considering the diversity and complexity of different makeup styles, makeup transfer remains a challenging task.

\begin{figure*}[t]
\centering
\includegraphics[width=0.9\linewidth]{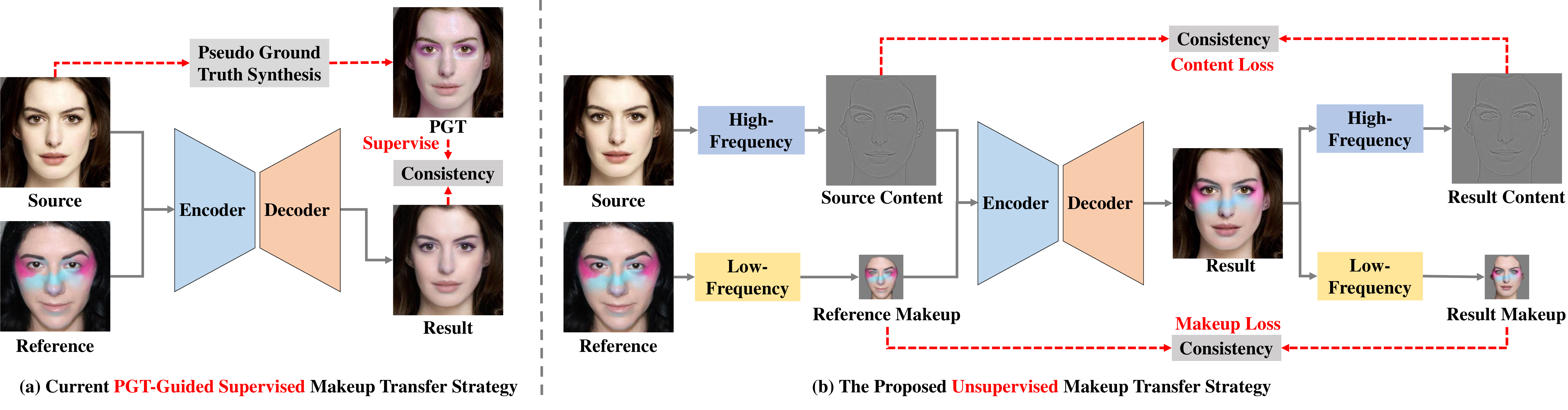}
\caption{The comparison of different training strategies.}
\label{img:introduction_training}
\end{figure*}

\begin{figure*}[t]
	\centering
	\includegraphics[width=1.0\linewidth]{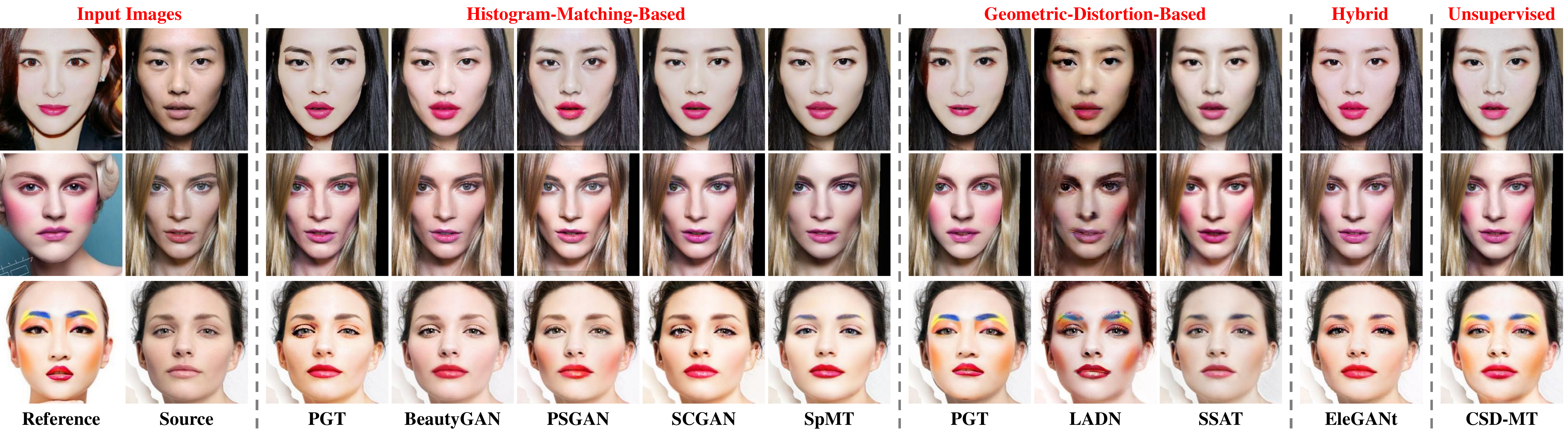}
	\caption{The PGTs and transferred results generated by different categories of makeup transfer methods.}
	\label{img:introduction_method}
\end{figure*}

One of the major problems with the makeup transfer task is its unsupervised nature, which means that there is no real transferred image that can be used as a target ground truth to guide the model training. To address this problem, most existing methods \cite{BeautyGAN,DMT,LADN,CPM,FAT,PSGAN,SCGAN,SpMT,SSAT,EleGANt} propose to synthesize a pseudo ground truth (PGT) image, from each input source-reference image pair, as an alternative supervision target. After that, the model parameters are optimized by minimizing the difference between each generated transferred output and its corresponding PGT (see Figure \ref{img:introduction_training}(a)).

As illustrated in Figure \ref{img:introduction_method}, according to the PGT generation strategy used, previous makeup transfer approaches can be roughly divided into two categories: (1) \textbf{Histogram-matching-based methods} \cite{BeautyGAN, PSGAN, PSGAN++, SCGAN, SpMT} attempt to align the color distribution of each facial region (e.g., lip, eye and face areas) in the source image with that of the same region in the reference face. However, the PGT produced by this strategy discards all spatial information of makeup styles, and usually suffers from the large color difference (e.g., eye shadow and powder blusher) between the source and reference images. (2) \textbf{Geometric-distortion-based methods} \cite{LADN, SSAT} synthesize the PGT by warping the reference face so that its shape (typically represented by some facial landmarks) is matched to that of the source one. But such process often introduces undesired artifacts and also leads to the loss of source content information. As a result, these low-quality PGTs will consequently degrade the transfer performance of all above-mentioned methods. Although a recent effort \cite{EleGANt} has been made to create more effective PGTs through a hybrid strategy, the generated PGTs are still sub-optimal and their imprecision will severely affect the final transferred results.

To eliminate these negative effects of the PGT, in this paper, we propose a new Content-Style Decoupled Makeup Transfer (CSD-MT) method, which works in a purely unsupervised manner without generating any PGT. To achieve this, one important observation is that the main differences of the same face image before and after makeup are concentrated in its low-frequency (LF) component, while the high-frequency (HF) component remains almost unchanged. Therefore, we can assume that \textit{the LF component of a face image is more associated with its makeup style information, and the HF component is more related to its content details.} With this assumption, CSD-MT first preforms frequency decomposition on each input and output image to decouple their contents and makeup styles. Then, for model training, CSD-MT simultaneously maximizes the content and makeup consistencies of the transferred result with the source and reference images based on their HF and LF components, respectively, as shown in Figure \ref{img:introduction_training}(b). Additionally, we introduce two novel loss functions to enhance the transfer of the spatial and color information in makeup. The effectiveness of our proposed CSD-MT method is evaluated on three publicly available datasets, covering various makeup styles as well as different pose and expression variations. Our main contributions are summarized as follows:

\begin{itemize}
  \item We propose a novel Content-Style Decoupled Makeup Transfer (CSD-MT) method, which works in a purely unsupervised manner. Based on frequency decomposition, CSD-MT for the first time eliminates the requirement of generating pseudo ground truth.
  \item Two newly designed loss functions, namely the self-augmented reconstructive loss and the color contrastive loss, are introduced to facilitate a better transfer of the spatial and color information in makeup.
  \item Extensive quantitative and qualitative comparisons on three datasets indicate that CSD-MT outperforms seven state-of-the-art makeup transfer methods. In addition, the ablation study validates the superiority of our proposed unsupervised learning strategy over PGT-guided training.
\end{itemize}

%% file: sec/2_related_works.tex
\section{Related Works}
\subsection{Makeup Transfer}
During the past decade, makeup transfer has attracted increasing attention from the computer vision community.
According to the PGT generation strategy, the previous methods can be roughly divided into two categories.
(1) \textbf{For histogram-matching-based methods}, BeautyGAN \cite{BeautyGAN} pioneers a histogram matching loss and designs a dual input/output GAN to perform makeup transfer and removal simultaneously.
To handle misaligned head poses and facial expressions, SCGAN \cite{SCGAN} encodes component-wise makeup regions into spatially-invariant style codes, while PSGAN \cite{PSGAN, PSGAN++} utilizes an attention mechanism to adaptively deform the makeup feature maps based on source images.
CPM \cite{CPM} bypasses semantic alignment by converting the images to UV \cite{UV} space, where each pixel corresponds to a specific semantic point on the face.
RamGAN \cite{RamGAN} and SpMT \cite{SpMT} explore local attention to eliminate potential associations between different makeup components.
(2) \textbf{For geometric-distortion-based methods}, PairedCycleGAN \cite{PairedCycleGAN} trains a style discriminator to measure the makeup consistency between the results  and the reference images.
For instances of extreme makeup, LADN \cite{LADN} employs multiple overlapping local makeup style discriminators.
To ensure color fidelity, FAT \cite{FAT} and SSAT \cite{SSAT} utilize cross-attention \cite{Transformer} to calculate semantic correspondence between the two input images.
In addition, the recent EleGANt \cite{EleGANt} achieves a more effective PGT using a hybrid strategy that integrates the advantages of these two PGTs by dynamically assigning different weights.

Different from all the above methods, whose transfer performance is heavily influenced by the quality of PGTs, the goal of our CSD-MT is to investigate a PGT-free makeup transfer approach to eliminate the negative effects of generating PGTs.

\begin{figure}[t]
	\centering
	\includegraphics[width=0.9\linewidth]{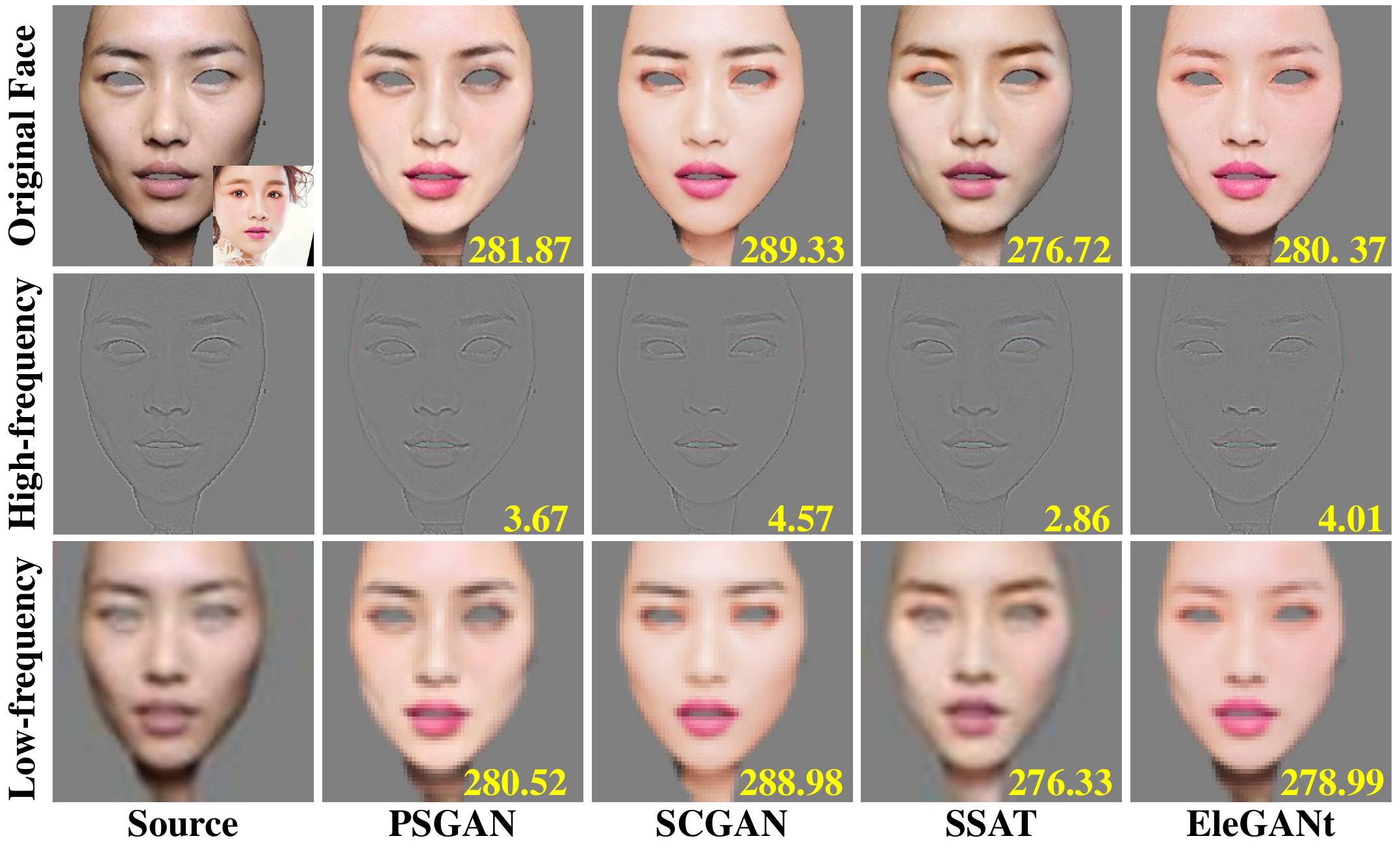} 
	\caption{ Visualization of the frequency components decomposed from the source image and the transferred results.
		The low-frequency components are resized for better visualization.
		The mean square errors of the different components between source images and transferred results are marked in the lower left corner.
	}
	\label{img:frequency_decomposition}
\end{figure}

\subsection{Frequency Decomposition}
Frequency decomposition has shown its effectiveness in various tasks, including classification \cite{LRR, WPCNN}, image synthesis \cite{LPGAN,LPSR}, and image translation \cite{WCT2, LPTN,ABPN}.
For instance, LRR \cite{LRR} utilizes the Laplacian pyramid \cite{LP} to refine the boundary details of semantic segmentation.
LapSRN \cite{LPSR} consists of multiple generators that progressively reconstruct the HF residuals of high-resolution images.
WTC$^{2}$ \cite{WCT2} employs wavelet transform to accelerate the stylization process of high-resolution images.
For makeup transfer, we observe that the main differences of the same face image before and after makeup are concentrated in its LF component, while the HF component remains almost unchanged.
Therefore, unlike the methods mentioned above, our goal of frequency decomposition is to decouple the content information and makeup style from a face image.

%% file: sec/3_methodology.tex
\section{Methodology}
\subsection{Problem Statement}
Let $\mathcal{X}$ and $\mathcal{Y}$ denote the non-makeup source domain and the makeup reference domain, respectively. In general, the image samples in $\mathcal{X}$ and $\mathcal{Y}$ are unpaired, which means that the source and reference images are collected from different persons with distinct identity information, and each reference face showcases a unique makeup style. Given a pair of source and reference images $\{(x,y)|x\in\mathcal{X}, y\in\mathcal{Y}\}$ as input, the main goal of makeup transfer is to generate a transferred result $\hat{x}$, which maximally preserves the content information in ${x}$ and contains the same makeup style as ${y}$. Such task can be considered as a cross-domain image-to-image translation problem with specific conditions, while its unsupervised nature makes it even more challenging.

\begin{figure*}[t]
\centering
\includegraphics[width=0.9\linewidth]{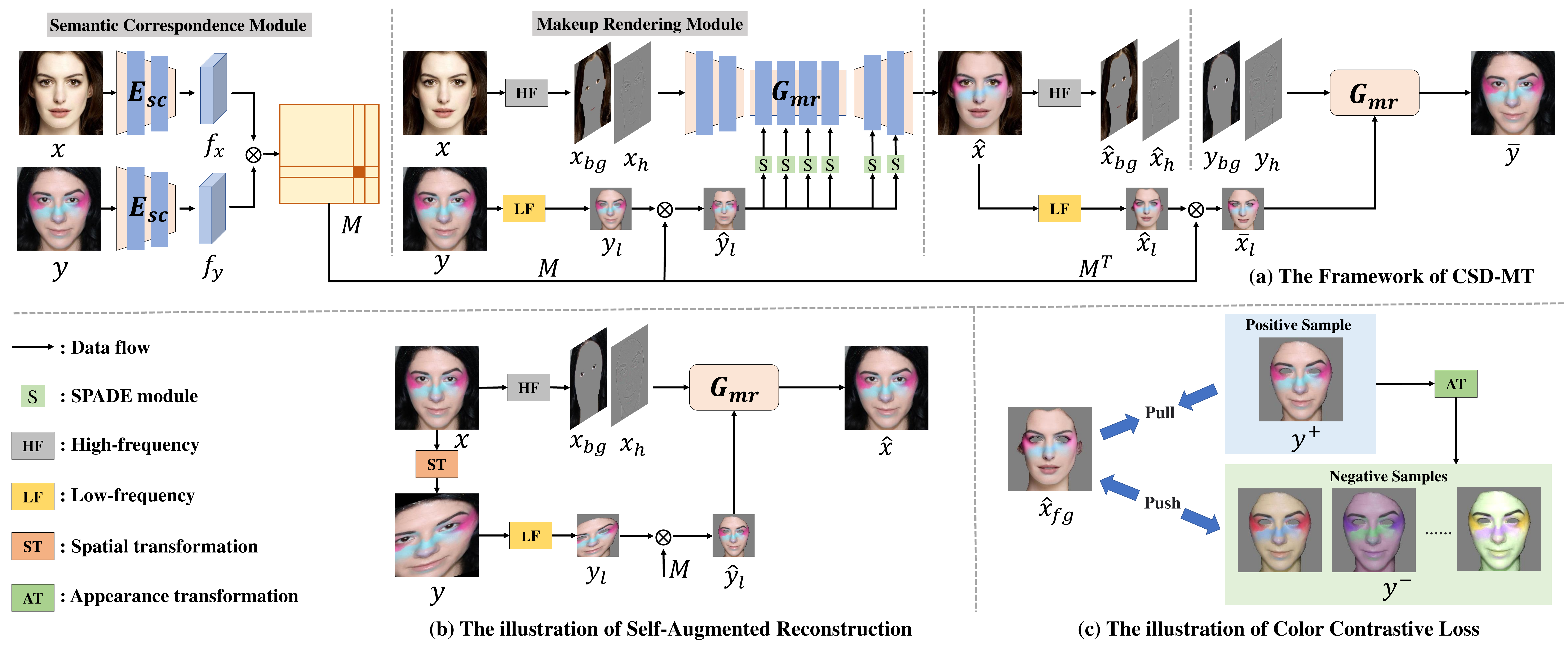} 
\caption{
Illustration of the proposed CSD-MT framework.
(a) Given a source image $x$ and a reference image $y$, the semantic correspondence module first constructs a pixel-wise correlation matrix $M$ between them.
Next, by performing face parsing and frequency decomposition, the makeup rendering module $G_{mr}$ obtains the background area $x_{bg}$ and the HF component $x_{h}$ that contain the content information of $x$, as well as the LF component $y_{l}$ comprising the makeup style of $y$.
Then, each pixel in $\hat{y}_{l}$ aggregates the information from the corresponding pixels in $y_{l}$ according to the correlation matrix $M$.
Finally, the final transferred result $\hat{x}=G_{mr}([x_{bg}$, $x_{h}],\hat{y}_{l})$ is generated using $x_{bg}$, $x_{h}$ and $\hat{y}_{l}$.
Furthermore, we introduce a self-augmented reconstructive loss (b) and a color contrastive loss (c) to enhance the transfer of the spatial and color information in makeup, respectively.
}
\label{img:network}
\end{figure*}

\subsection{Content and Makeup Style Decoupling}
In order to solve the makeup transfer task without generating PGT, we attempt to seek the content and makeup style supervision signals from the input images $(x,y)$ themselves to accurately control the corresponding information in $\hat{x}$. We approach this purpose by investigating the frequency characteristics of these two types of information. To do this, we randomly select 500 pairs of test images from the MT dataset \cite{BeautyGAN}, and perform frequency decomposition on each source image and its corresponding transferred results generated by different methods. More specifically, given an arbitrary image $x\in\mathbb{R}^{H\times W\times 3}$ where $H$ and $W$ denote its height and width, we first remove its background region $x_{bg}$ through a face parsing technique \cite{BiSeNet}. After that, by applying a fixed Gaussian kernel on the remaining foreground face image $x_{fg}$, we calculate a low-pass prediction $x_{l}\in \mathbb{R}^{\frac{H}{d}\times\frac{W}{d}\times 3}$, where $d$ represents a downsampling factor. Based on this, the high-frequency residual $x_{h}\in\mathbb{R}^{H\times W\times 3}$ is finally obtained by $x_{h}=x_{fg}-up(x_{l})$, where $up(\cdot)$ is a bilinear interpolation upsampling operation.

We measure the mean squared errors (MSE) of these decomposed LF and HF components between source and transferred images in Figure \ref{img:frequency_decomposition}. It can be observed that the MSE values calculated on the LF components are much larger than those obtained on the HF components. This suggests that the main differences of the same face images before and after makeup are primarily concentrated in their LF components, while the HF components remain almost unchanged. Additionally, the visual comparisons displayed in Figure \ref{img:frequency_decomposition} also support this claim. Therefore, we can assume that \textit{the LF component of a face image is more associated with its makeup style information, and the HF component is more related to its content details.} Such assumption allows us to appropriately decouple the content information and makeup style contained in each face image by using the frequency decomposition process described above.

\subsection{The Proposed CSD-MT Method}
To highlight that the improvement of transfer performance mainly comes from our proposed unsupervised learning strategy, we design a relative concise architecture for the proposed CSD-MT method. As illustrated in Figure \ref{img:network}, the generator $\mathcal{G}$ of CSD-MT contains a semantic correspondence module and a makeup rendering module, which are presented in detail in the following subsections.

\noindent \textbf{Semantic Correspondence Module.} Generally, due to the differences in head pose and facial expression, the same facial parts in the input source image $x$ and reference image $y$ often appear at different spatial locations \cite{PSGAN,SSAT,EleGANt}, and such semantic misalignment will eventually lead to performance degradation. To alleviate this problem, our semantic correspondence module constructs a pixel-wise correlation matrix $M$ by calculating the cosine similarity as:
\begin{equation}
	\label{equ1}
	\begin{aligned}
		M(i,j)=\frac{f_{x}(i)^T f_{y}(j)}{\|f_{x}(i)\|_{2}\|f_{y}(j)\|_{2}}. \\
	\end{aligned}
\end{equation}
Here, $f_{x}=E_{sc}(x)$, $f_{y}=E_{sc}(y)$ denote the semantic features extracted by a convolutional encoder $E_{sc}(\cdot)$. Both $f_{x}$ and $f_{y}$ have the same spatial resolution as the LF component of input images, i.e., $\frac{H}{d}\times\frac{W}{d}$. $f(i)$ represents the feature vector of the $i$-th pixel in $f$ and $M(i,j)$ indicates the element at the $(i,j)$-th location of $M$. We consider the correlation matrix $M$ as a deformation mapping function, and use it to achieve semantic alignment between the source and reference images in our makeup rendering module.

\begin{figure*}[t]
	\centering
	\includegraphics[width=0.9\linewidth]{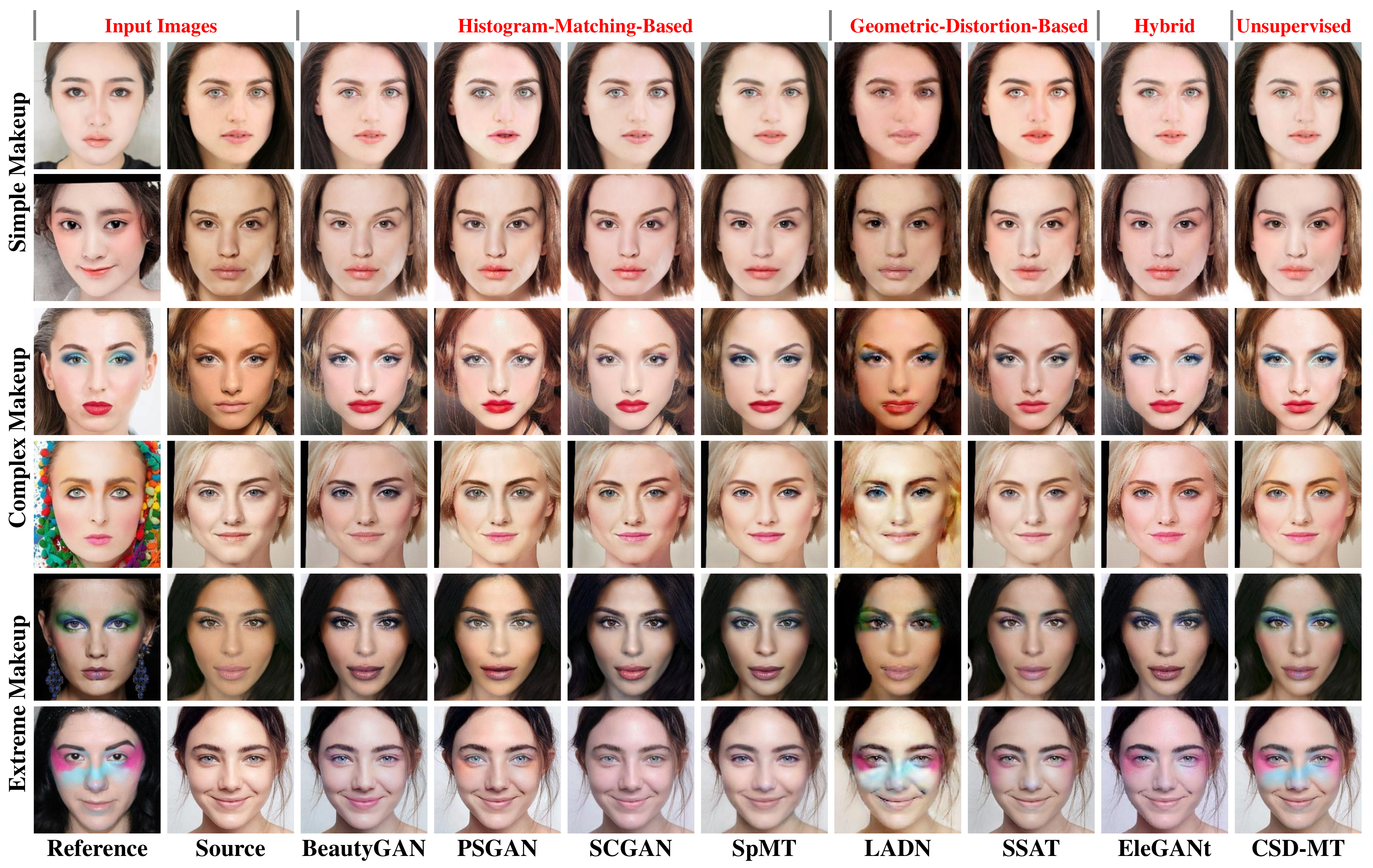} 
	\caption{
		Qualitative comparison with several state-of-the-art methods on different makeup styles. The proposed CSD-MT produces the most precise transferred results with desired makeup information and high-quality content details.
		Please zoom in for better comparison.
	}
	\label{img:qualitative_comparison}
\end{figure*}

\noindent \textbf{Makeup Rendering Module.} By performing face parsing and frequency decomposition on the input images, we can obtain the background area $x_{bg}$ and the HF component $x_{h}$ that contain the content information of the source image $x$, as well as the LF component $y_{l}$ comprising the makeup style of the reference image $y$. Then, the correlation matrix $M$ in Eq. (\ref{equ1}) is used to spatially deform $y_{l}$ as follow:
\begin{equation}
	\label{equ2}
	\begin{aligned}
		\hat{y}_{l}(i)=\sum\nolimits_{j}Softmax(M(i,j) / \tau)\cdot y_{l}(j),
	\end{aligned}
\end{equation}
where $Softmax(\cdot)$ denotes a softmax computation along the column dimension, which normalizes the element values in each row of $M$, and $\tau>0$ is a temperature parameter. Based on $M$, each pixel in $\hat{y}_{l}$ aggregates the information from the corresponding pixels in $y_{l}$ according to the semantic correspondence between $x$ and $y$. Therefore, the deformed $\hat{y}_{l}$ is semantically aligned with the source image.

Finally, the makeup rendering module generates the final transferred result $\hat{x}$ based on $x_{bg}$, $x_{h}$ and $\hat{y}_{l}$ as:
\begin{equation}
	\label{equ3}
	\begin{aligned}
		\hat{x}=G_{mr}([x_{bg}, x_{h}], \hat{y}_{l}).
	\end{aligned}
\end{equation}
Here, $[\cdot, \cdot]$ denotes a channel-wise concatenation operation. $G_{mr}(\cdot, \cdot)$ is an encoder-decoder network implemented with the U-Net structure. In $G_{mr}(\cdot, \cdot)$, the conditional makeup information is introduced through the SPADE modules \cite{SPADE}, whose modulation parameters are generated from $ \hat{y}_{l}$.

\subsection{Training Objectives}
\label{sec:loss}

\noindent \textbf{Transfer Loss.} Similar to the input images, the transferred result $\hat{x}$ produced by our generator $\mathcal{G}$ can also be decomposed into the LF component $\hat{x}_{l}$ and the HF component $\hat{x}_{h}$. By employing the transposed correlation matrix $M^T$ as in Eq. (\ref{equ2}), we re-deform $\hat{x}_{l}$ into $\bar{x}_{l}$ to make it semantically aligned with the reference image. According to our assumption, $\bar{x}_{l}$ is expected to be consistent with ${y}_{l}$, such that the makeup style can be faithfully transferred. And meanwhile, $\hat{x}_{h}$ is required to preserve the content information of the source image and thus should be consistent with $x_{h}$. Therefore, the following transfer loss is defined to simultaneously promote the makeup and content consistencies:
\begin{equation}
	\label{equ4}
	\begin{aligned}
    	L_{trans}=&L_{makeup}+\alpha L_{cont}, \\
    	L_{makeup}=\|\bar{x}_{l}&-y_{l}\|_{1}, L_{cont}=GP(\hat{x}_{h},x_{h}),
	\end{aligned}
\end{equation}
where $\alpha>0$ balances the importance of the two terms. $L_{makeup}$ is defined as the L1 distance between $\bar{x}_{l}$ and $y_{l}$. $L_{cont}$ calculates the Gradient Profile loss \cite{SPL} $GP(\cdot,\cdot)$ between $\hat{x}_{h}$ and $x_{h}$, which is computed in the image gradients space and thus is more powerful in distilling HF details.

\noindent \textbf{Cycle Consistency Loss.}
Inspired by CycleGAN \cite{CycleGAN}, we feed $y_{bg}, y_{h}$ and $\bar{x}_{l}$ into the makeup rendering network in Eq. (\ref{equ3}) once again. We expect the produced transferred result $\bar{y}=G_{mr}([y_{bg}, y_{h}], \bar{x}_{l})$ can be as close as possible to the original reference image $y$, which can be formulated as:
\begin{equation}
	\label{equ6}
	\begin{aligned}
		L_{cycle}=\|\bar{y}-y\|_{1}.
	\end{aligned}
\end{equation}

\noindent \textbf{Adversarial Loss.} To make the transferred results more realistic, we construct a multi-scale discriminator $\mathcal{D}$ \cite{Pix2PixHD} to distinguish the face images generated by CSD-MT from the reference images containing real makeup information. Based on objective function of LSGAN \cite{LSGAN}, our adversarial loss $L_{adv}$ is defined as follow:
\begin{equation}
\label{equ7}
  L_{adv}=\mathbb{E}_{y}[(\mathcal{D}(y)-1)^{2}]+ \mathbb{E}_{\hat{x}}[(\mathcal{D}(\hat{x}))^{2}].
\end{equation}

\begin{table*}[t]\small
	\begin{center}
		\begin{tabular}{l|c|c|c|c|c|c|c|c|c}
			\hline
			\multirow{3}{*}{Methods} & \multicolumn{3}{c|}{MT}& \multicolumn{3}{c|}{Wild-MT} & \multicolumn{3}{c}{LADN} \\
			\cline{2-4} \cline{5-7} \cline{8-10}
			
			& \multirow{2}{*}{FID} & \multicolumn{2}{c|}{Self-Aug PSNR/SSIM} & \multirow{2}{*}{FID} & \multicolumn{2}{c|}{Self-Aug PSNR/SSIM}& \multirow{2}{*}{FID} & \multicolumn{2}{c}{Self-Aug PSNR/SSIM}  \\
			\cline{3-4} \cline{6-7} \cline{9-10}
			&  & Crop & Rotate   & & Crop & Rotate & & Crop & Rotate  \\
			\hline
			BeautyGAN \cite{BeautyGAN}  &48.24 & 21.70/0.848& 21.58/0.847 & 99.62&21.67/0.875 &21.33/0.862 & 69.50&19.73/0.719 &20.39/0.731 \\
			PSGAN \cite{PSGAN} &45.02 & 23.46/0.886&22.68/0.875  & 89.92&22.56/0.879&21.89/0.870 & 57.80&22.97/0.824 &22.16/0.815 \\
			SCGAN \cite{SCGAN} &39.20 &23.92/0.887&24.05/0.888  & 79.54&24.02/0.902 &24.24/0.905 & 51.39&22.42/0.839 &22.44/0.838\\
			SpMT \cite{SpMT} &46.10 &24.36/0.891&23.79/0.887  & 77.04&24.30/0.909 &23.77/0.904 & 48.18&23.31/0.817 &22.89/0.815\\
			LADN \cite{LADN} &73.91 & 19.27/0.759&19.02/0.755  & 104.91&19.22/0.790 &19.09/0.786 & 65.87&18.61/0.727 &18.39/0.723\\
			SSAT \cite{SSAT} &38.01 & 24.01/0.894&23.93/0.893  & 70.53&23.57/0.905 &23.80/0.907  & 53.84&22.81/0.848 &22.95/0.846\\
			EleGANt \cite{EleGANt} &54.06 & 25.15/0.885&24.60/0.875  & 86.19&24.82/0.893 &24.49/0.877 & 61.40&24.64/0.838 &24.31/0.848\\
			CSD-MT (Ours) &\textbf{37.56} & \textbf{27.28/0.920}&\textbf{26.68/0.915} & \textbf{60.82}&\textbf{27.83/0.934} &\textbf{26.70/0.923} & \textbf{40.87}&\textbf{25.19/0.868} &\textbf{25.23/0.868} \\
			\hline
		\end{tabular}
	\end{center}
	\caption{Quantitative comparison of FID and Self-augmented PSNR/SSIM on the MT, Wild-MT and LADN datasets.}
	\label{table:SSIM_FID}
\end{table*}

\noindent \textbf{Self-Augmented Reconstructive Loss.} To further enhance the robustness to different head poses and facial expressions, we develop a self-augmented reconstruction process that facilitates a better transfer of the spatial information in makeup, as shown in Figure \ref{img:network}(b). Specifically, given an image $x$ with makeup, we impose a random spatial transformation $T_s(\cdot)$ on it and obtain a transformed image $y=T_s(x)$. This image has the same makeup style as $x$, but the spatial information in the original $x$ (e.g., facial structure and person identity) is completely destroyed. Considering $x$ as the source image and $y$ as the reference image, the generator $\mathcal{G}$ of CSD-MT takes $(x,y)$ as input and outputs a transferred result $\hat{x}$. Based on this process, $\hat{x}$ should faithfully reconstruct $x$, since it contains the same content and makeup style (distilled from $y$) information as in $x$. Therefore, we define the following self-augmented reconstructive loss:
\begin{equation}
\label{equ8}
\begin{aligned}
	L_{aug}=\|\hat{x}-x\|_{1}.
\end{aligned}
\end{equation}

\noindent \textbf{Color Contrastive Loss.} To promote the color fidelity, as shown in Figure \ref{img:network}(c), we propose a color contrastive loss which can be formulated as follow:
\begin{equation}
	\label{equ9}
	\begin{aligned}
		L_{cts}=-log(1-\frac{\ell(\hat{x}_{fg},y^{+})}{\sum\nolimits_{i=1}^{N}\ell(\hat{x}_{fg},y_{i}^{-})}),
	\end{aligned}
\end{equation}
where $\hat{x}_{fg}$ is the foreground face area separated from the transferred result $\hat{x}$. For this anchor, $y^{+}$ and $y^{-}$ denote the positive and negative samples, respectively. In our implementation, the face area $y_{fg}$ of the input reference image is used as the only positive sample. And each negative sample $y_{i}^{-}$ is obtained by performing a random appearance transformation $T_a(\cdot)$ on $y_{fg}$, i.e., $y_{i}^{-}=T_a(y_{fg})$, $N$ is the total number of negative samples. Based on the perceptual loss \cite{Perceptual}, the distance function $\ell(\cdot,\cdot)$ is defined as:
\begin{equation}
\label{equ10}
\begin{aligned}
  \ell(x,y)=\sum\nolimits_{l}\|Gram(\phi_{l}(x))-Gram(\phi_{l}(y))\|_{1},
\end{aligned}
\end{equation}
where $\phi_{l}(\cdot)$ represents the feature map extracted from the $l$-th layer of the pre-trained VGG19 \cite{VGG} model. $Gram(\cdot)$ calculates the gram matrix of a feature map. By minimizing the color contrastive loss in Eq. (\ref{equ9}), $\hat{x}_{fg}$ and $y_{fg}$ with similar color distributions are pulled closer, while $\hat{x}_{fg}$ and $y_{i}^{-}$ with different color distributions are pushed away.

\noindent \textbf{Overall Loss.}
In summary, the overall loss function for the generator $\mathcal{G}$ and discriminator $\mathcal{D}$ of the proposed CSD-MT method is defined as:
\begin{equation}
\label{equ11}
\begin{aligned}
  \min_{\mathcal{G}}\max_{\mathcal{D}}L&=\lambda_{trans}L_{trans}+\lambda_{cycle}L_{cycle}\\
                 &+\lambda_{adv}L_{adv}+\lambda_{aug}L_{aug}+\lambda_{cts}L_{cts}.
\end{aligned}
\end{equation}

%% file: sec/4_experiment.tex
\begin{figure*}[t]
	\centering
	\includegraphics[width=0.7\linewidth]{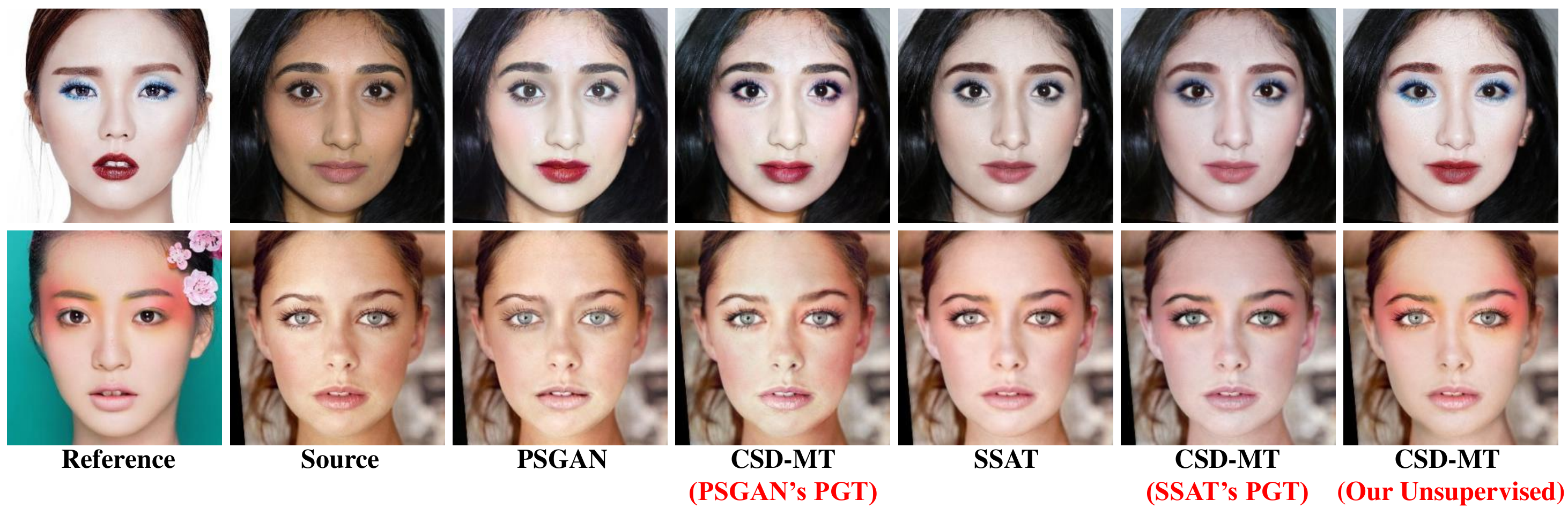} 
	\caption{
		Transferred results produced by the CSD-MT models trained with different learning strategies.
	}
	\label{img:PGTvsUnsupervised}
\end{figure*}

\section{Experiments}
\subsection{Datasets}
\textbf{MT Dataset} \cite{BeautyGAN} contains 1,115 non-makeup and 2,719 makeup images, which are mostly well-aligned and have plenty of makeup styles. We split the training and testing sets by following the strategy in \cite{BeautyGAN,PSGAN}.

\noindent\textbf{Wild-MT Dataset} \cite{PSGAN} consists of 369 non-makeup and 403 makeup images. Most of them contain large variations in head pose and facial expression.

\noindent\textbf{LADN Dataset} \cite{LADN} has 333 non-makeup and 302 makeup images, including 155 extreme makeup images with great variances on makeup color, style and region coverage.

\subsection{Baselines}
We compare our proposed CSD-MT approach with seven state-of-the-art makeup transfer methods, including four histogram-matching-based methods (BeautyGAN \cite{BeautyGAN},  PSGAN \cite{PSGAN}, SCGAN \cite{SCGAN}, and SpMT \cite{SpMT}), two geometric-distortion-based methods (LADN \cite{LADN}, SSAT \cite{SSAT}), and one hybrid method (EleGANt \cite{EleGANt}).

All these methods are trained by only using the training set of the MT dataset. And their performance and generalization ability are evaluated on the test set of the MT dataset, as well as on the Wild-MT and LADN datasets. See supplementary materials for the training details of CSD-MT.


\begin{figure}[t]
	\centering
	\includegraphics[width=0.75\linewidth]{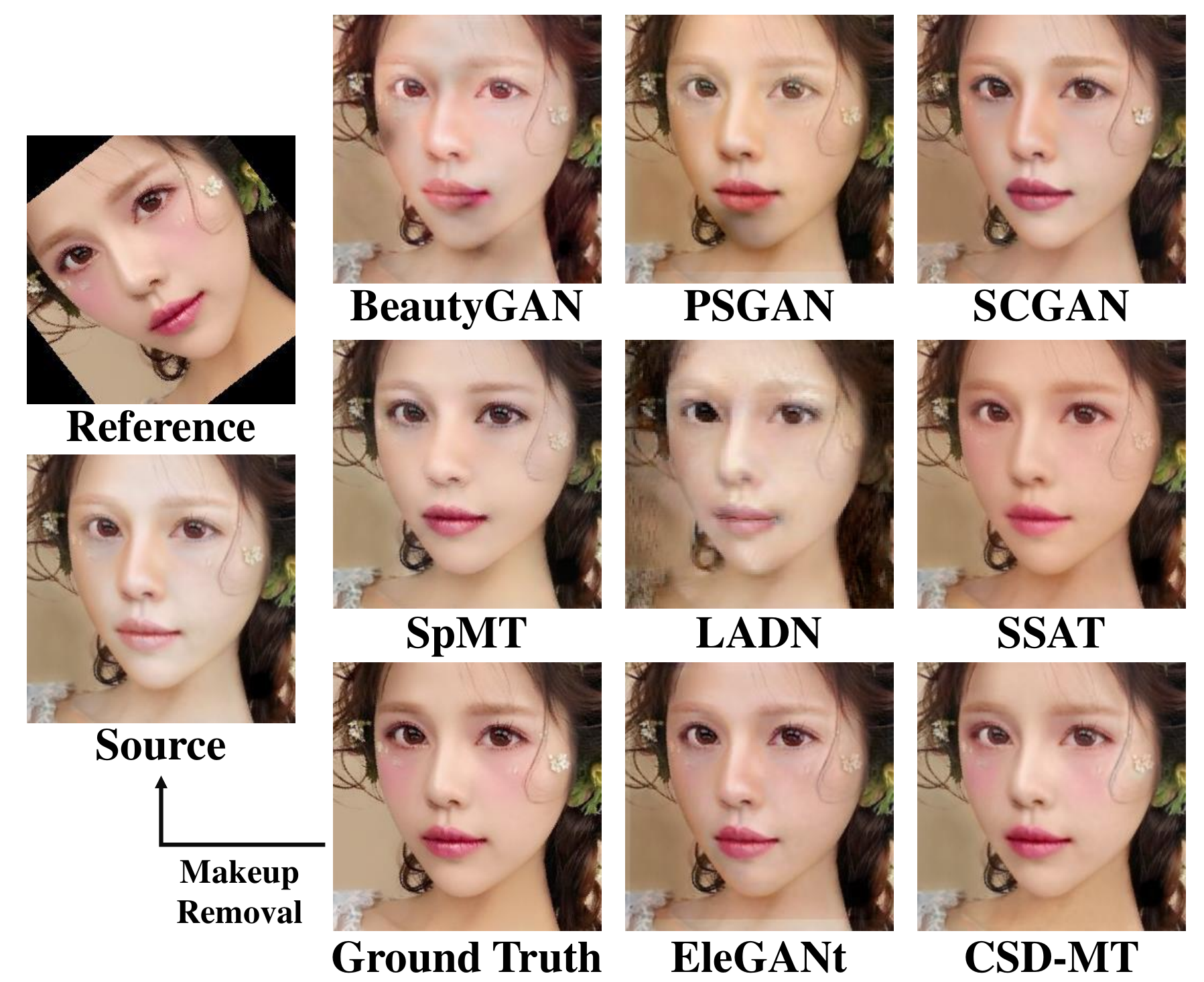}
	\caption{
		Reconstruction results obtained by rendering the de-makeup source image using a randomly rotated reference image.
	}
	\label{img:augmented-self_reconstruction}
\end{figure}

\begin{table}[t]\small
	\begin{center}
		\begin{tabular}{l|c|c|c}
			\hline
			Methods & Simple & Complex & Extreme \\
			\hline
			SCGAN \cite{SCGAN} & 5.4\% & 1.7\% & 1.1\% \\
			SpMT \cite{SpMT} & 11.1\% & 2.6\% & 1.5\% \\
			SSAT \cite{SSAT} & 13.3\% & 9.8\% & 7.4\% \\
			EleGANt \cite{EleGANt} & 14.3\% & 15.4\% & 13.2\% \\
			CSD-MT (Ours) & \textbf{55.9\%} & \textbf{70.4\%} & \textbf{76.7\%} \\			
			\hline
		\end{tabular}
	\end{center}
	\caption{The ratio selected as best (\%) on different types of makeup styles.
		We classify "Simple", "Complex" and "Extreme" makeup based on our subjective experience.}
	\label{tabl:user_study}
\end{table}

\subsection{Qualitative Comparison}
Figure \ref{img:qualitative_comparison} displays the transferred results of all competing methods on various makeup styles. It can be seen that the histogram-matching-based methods all fail to work when there are large color differences between the source and reference images. For geometric-distortion-based methods, LADN generates unrealistic results with undesired artifacts, while SSAT cannot effectively transfer the makeup details such as the eye shadow and lipstick. The hybrid method EleGANt achieves better results than other baselines, but still struggles with transferring extreme makeup styles that distributed throughout the entire facial area. In contrast, for all types of makeup styles, our unsupervised CSD-MT method produces the most precise transferred results with desired makeup information and high-quality content details.

\subsection{Quantitative Comparison}

\noindent \textbf{Fr\'echet Inception Distance (FID).}
Following \cite{PSGAN++}, we calculate the FID score \cite{FID} (lower is better) between the reference images and the transferred results generated by different methods, which are reported in Table \ref{table:SSIM_FID}. The lowest FID scores achieved by our CSD-MT method indicate that its outputs are more realistic.

\begin{table}[t]\small
\begin{center}
\begin{tabular}{l|c|c}
        \hline
        \multirow{2}{*}{Methods} & \multicolumn{2}{c}{Self-Aug PSNR/SSIM}  \\
        \cline{2-3}
      &  Crop & Rotate \\
        \hline
         CSD-MT with PSGAN's PGT   & 22.22/0.881 & 22.37/0.883 \\
         CSD-MT with SSAT's PGT   & 23.23/0.893 & 23.37/0.891 \\
         \hline
         PSGAN with $L_{aug}$ and $L_{cts}$   & 23.17/0.883 & 22.42/0.874 \\
         SSAT with $L_{aug}$ and $L_{cts}$   & 23.78/0.894 & 23.64/0.891 \\
         \hline
         (Setting A) $L_{adv}+L_{cont}$  & 19.97/0.866 & 20.01/0.867 \\
         (Setting B) A+$L_{makeup}$   & 23.01/0.872 & 22.77/0.871 \\
         (Setting C) B+$L_{cycle}$   & 24.92/0.883 & 24.61/0.882 \\
         (Setting D) C+$L_{aug}$   & 26.45/0.914 & 25.61/0.907 \\
         (Setting E) D+$L_{cts}$   & \textbf{27.28/0.920} & \textbf{26.68/0.915} \\
        \hline
\end{tabular}
\end{center}
\caption{Quantitative results of ablation studies on the MT dataset.}
\label{tabl:ablation}
\end{table}

\begin{figure}[t]
	\centering
	\includegraphics[width=1.0\linewidth]{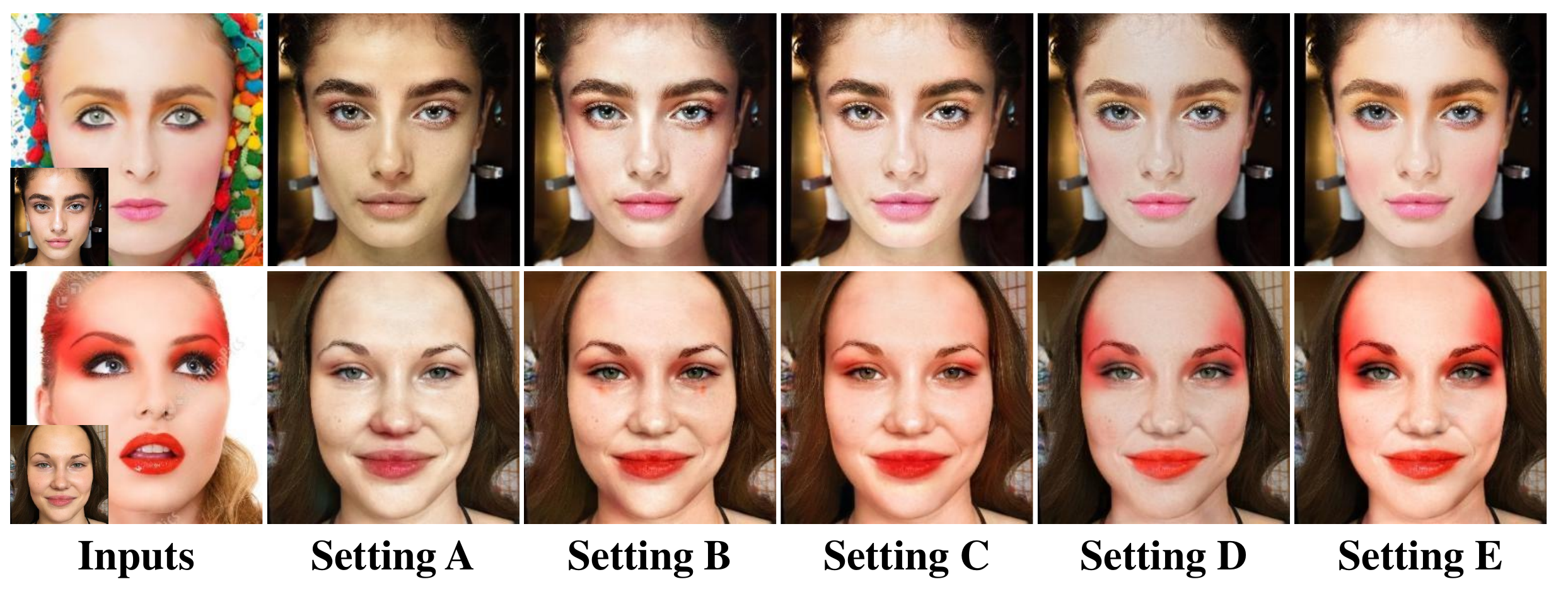}
	\caption{
		Transferred results produced by the CSD-MT models trained with different loss functions.
	}
	\label{img:effectOFloss}
\end{figure}

\noindent \textbf{Self-Augmented PSNR/SSIM.}
A major challenge in the performance evaluation for makeup transfer tasks is the lack of ground truth images. We utilize a similar self-augmented reconstruction process mentioned in section \ref{sec:loss} to address this issue. As shown in Figure \ref{img:augmented-self_reconstruction}, given a makeup sample, we randomly crop or rotate it to obtain a pseudo reference image whose content information has been corrupted. We also generate a de-makeup pseudo source image using the makeup removal function of SSAT. With these pseudo inputs, we treat the original makeup image as the ground-truth, and compute the PSNR and SSIM (higher is better) for model evaluation. Both the quantitative and qualitative results in Table \ref{table:SSIM_FID} and Figure \ref{img:augmented-self_reconstruction} show that CSD-MT outperforms other state-of-the-art methods, demonstrating its effectiveness in generating high-quality results.

\begin{figure}[t]
	\centering
	\includegraphics[width=1.0\linewidth]{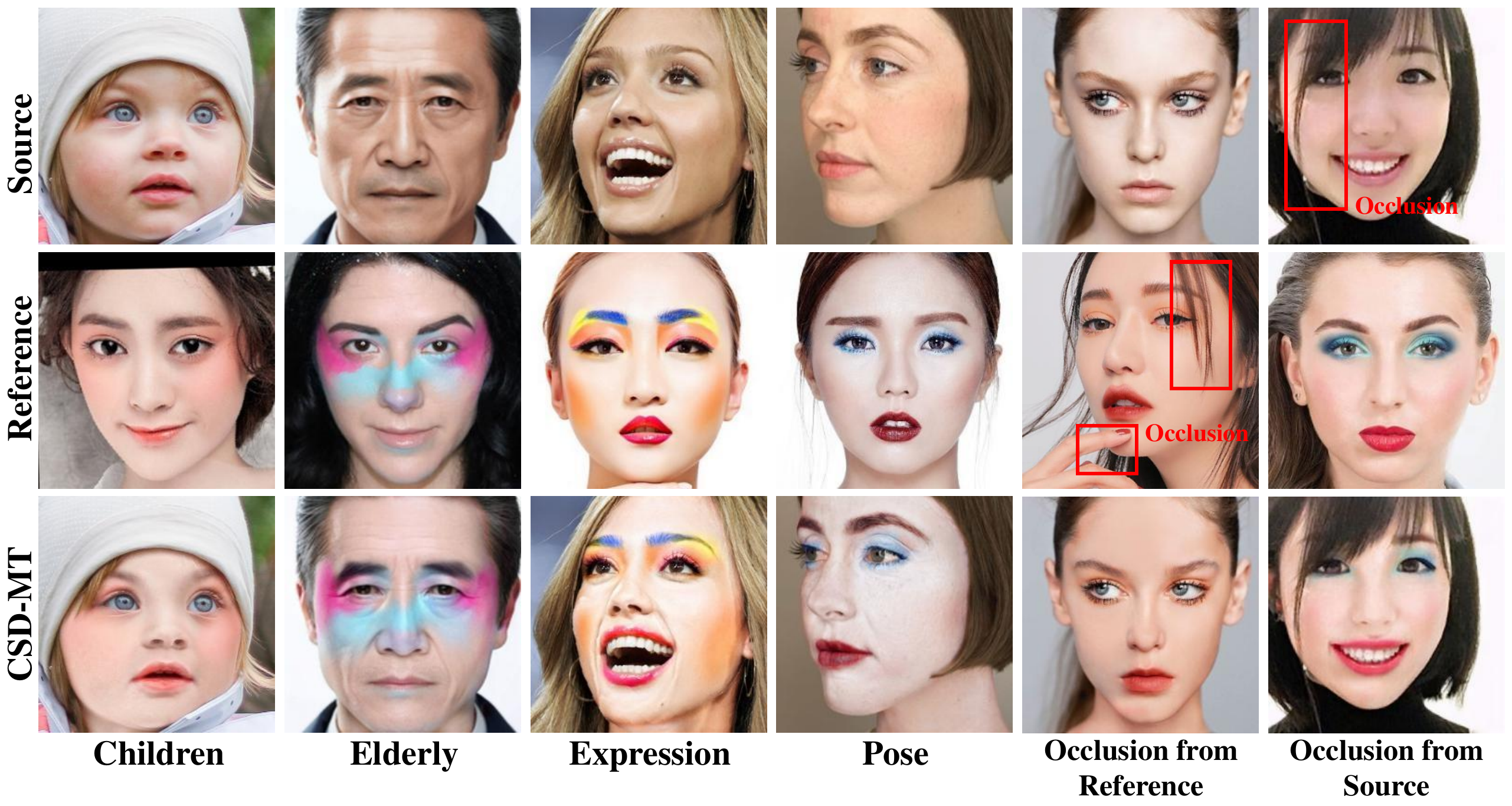}
	\caption{
		Robustness of CSD-MT in various complex scenarios.
	}
	\label{img:robust}
\end{figure}

\noindent \textbf{User Study.}
We also conduct a user study to quantitatively evaluate the performance of different models. We randomly select 20 pairs of images with different types of makeup styles and generate the transferred results using different methods. Then, totally 23 participants are asked to choose the most satisfactory result based on the image quality and makeup similarity. For a fair comparison, the transferred results are shown simultaneously in a random order. The results of user study are shown in Table \ref{tabl:user_study}.

\subsection{Ablation Study}
\noindent \textbf{PGT-guided vs. Unsupervised.} To intuitively compare these two strategies, we keep the network architecture of CSD-MT unchanged and train this model by using the PGT-guided training process as in PSGAN and SSAT, respectively. From Table \ref{tabl:ablation} and Figure \ref{img:PGTvsUnsupervised}, we can see that the outputs produced by the resulted models are very similar to those of PSGAN and SSAT. This suggests that the transfer performance is heavily influenced by the synthesized PGTs rather than the network structure. It can be also found that the original CSD-MT significantly outperforms these two models, demonstrating the superiority of our proposed unsupervised strategy over PGT-guided training.
Additionally, we integrate the losses $L_{aug}$ and $L_{cts}$ into PSGAN and SSAT. As shown in Table \ref{tabl:ablation}, the influence of $L_{aug}$ and $L_{cts}$ on previous methods is slight, indicating that our unsupervised strategy fits better with these two losses.

\noindent \textbf{Loss Functions}. As shown in Table \ref{tabl:ablation}, to analyze the effect of different losses defined in section \ref{sec:loss}, we gradually add each loss into a basic setting ($L_{adv}+L_{cont}$), resulting in 5 different loss combinations (Setting A-E). The quantitative and qualitative results of the CSD-MT models trained with these settings are displayed in Table \ref{tabl:ablation} and Figure \ref{img:effectOFloss}, respectively. We can observed that the model trained with only $L_{adv}+L_{cont}$ can already preserve the content details effectively. This is mainly attributed to the content-style decoupling operation and the content consistency loss $L_{cont}$. By adding $L_{makeup}$ and $L_{cycle}$, the makeup style information is transferred but some complex details are still missing. Further equipped with $L_{aug}$ facilitates the transfer of spatial information, so the makeup can appear at the correct location on the face. Finally, $L_{cts}$ ensures the color fidelity.

\noindent \textbf{Robustness.}
As shown in Figure \ref{img:robust}, capturing the semantic correspondence between the source and reference images makes our CSD-MT model insensitive to age, pose and expression variations. In addition, separating the foreground and background areas through the face parsing allows our method to be unaffected by image occlusions.

\begin{figure}[t]
	\centering
	\includegraphics[width=1.0\linewidth]{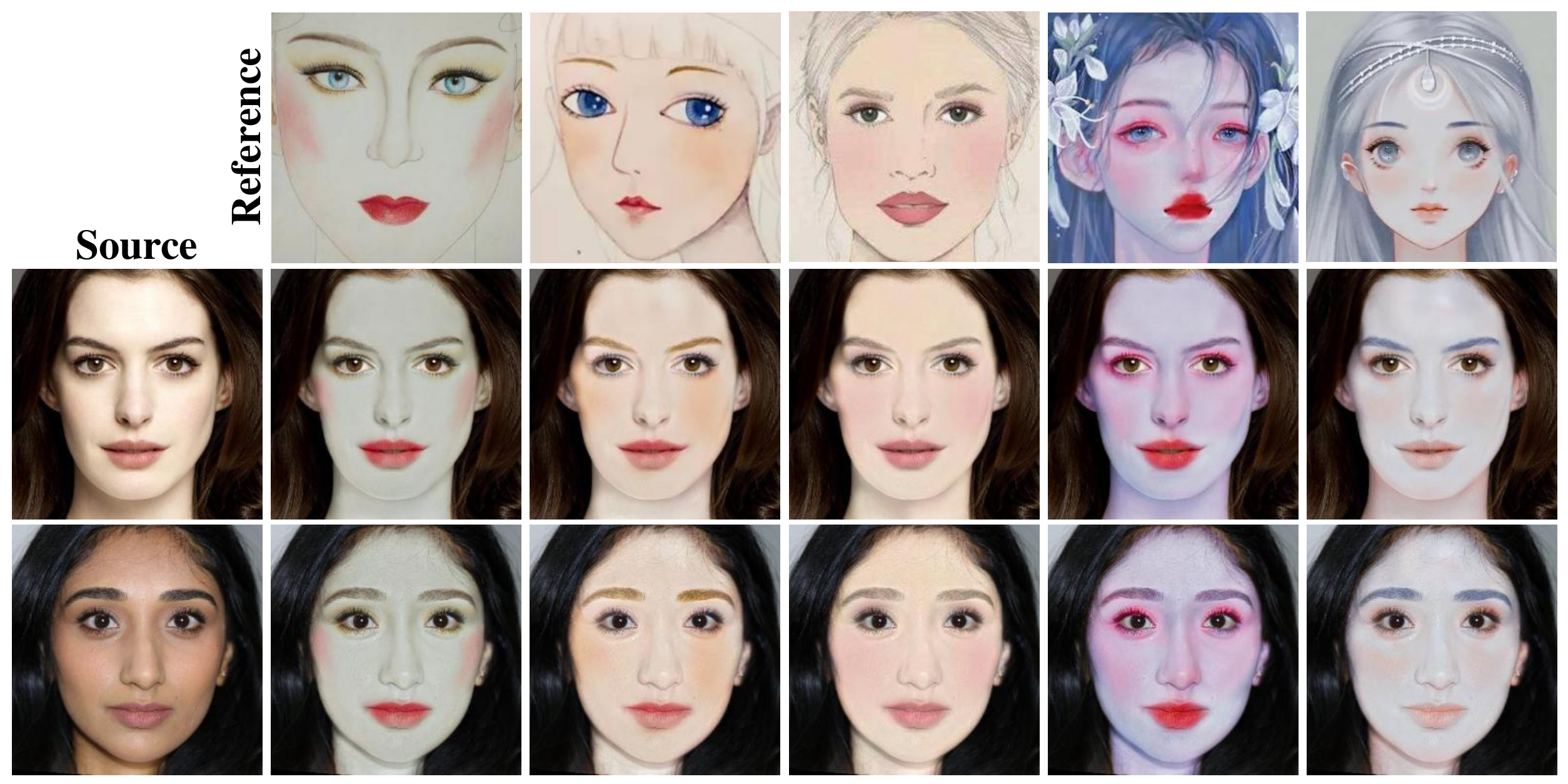}
	\caption{
		Generalization of CSD-MT in unseen anime makeup.
	}
	\label{img:generalization}
\end{figure}

\noindent \textbf{Generalization.}
To evaluate the generalization ability of our method, we collect some anime makeup examples which have a significant domain gap with the training samples in the MT dataset and have never been encountered by the model before.
The results are displayed in Figure \ref{img:generalization}.

\noindent \textbf{Limitation.}
In CSD-MT, we assume that the HF component is more closely associated with the content details of face images. This assumption may result in CSD-MT being unable to handle cases with patterns.

%% file: sec/5_conclusion.tex
\section{Conclusion}
In this paper, we propose an unsupervised makeup transfer method called CSD-MT to eliminate the negative effects of generating PGTs.
Inspired by the observed frequency characteristics, CSD-MT decouples the content and makeup style information through frequency decomposition and realizes makeup transfer by maximizing the consistency of these two types of information between the transferred result and input images, respectively.
Experiments demonstrate that our CSD-MT method significantly outperforms existing state-of-the-art methods in quantitative and qualitative analyses.

\noindent\textbf{Acknowledgments.}
This work was supported in part by the National Key Research and Development Program of China (Grant No.2022ZD0160604), the Project of Sanya Yazhou Bay Science and Technology City (Grant No.SCKJ-JYRC-2022-76 and SKJC-2022-PTDX-031), the Young Scientists Fund of the National Natural Science Foundation of China (Grant No.62306219), and the CAAl-Huawei MindSpore Open Fund (Grant No.CAAIXSILJJ-2022-004A).

%% file: sec/X_suppl.tex
\clearpage
\setcounter{page}{1}
\maketitlesupplementary

\section{Network Structure}
\begin{figure}[t]
	\centering
	\includegraphics[width=1.0\linewidth]{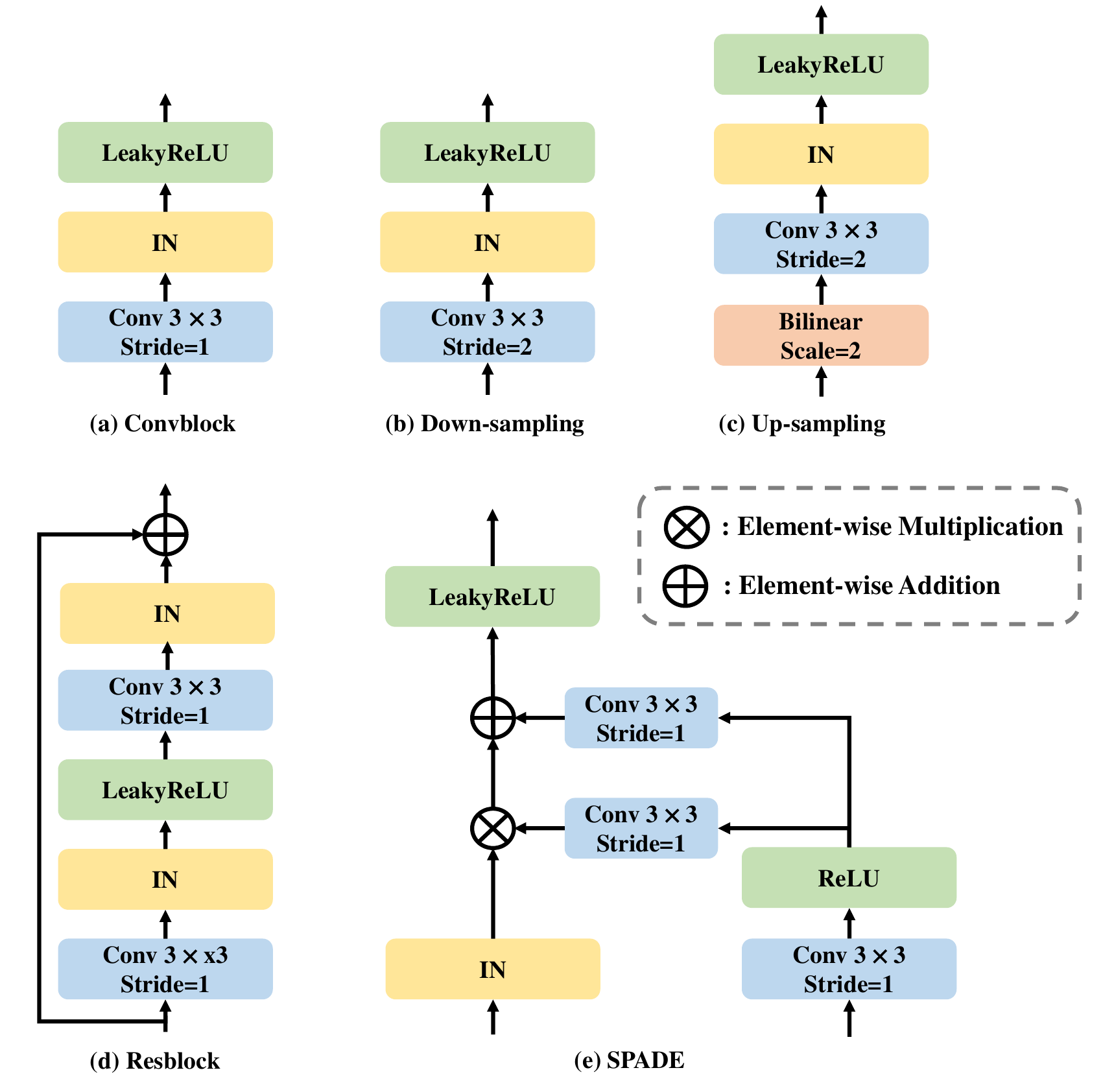}
	\caption{
		Basic network blocks used in the proposed CSD-MT model. Here, "IN" denotes an instance normalization layer.
	}
	\label{img:supply_basic_blocks}
\end{figure}

\begin{figure*}[t]
	\centering
	\includegraphics[width=1.0\linewidth]{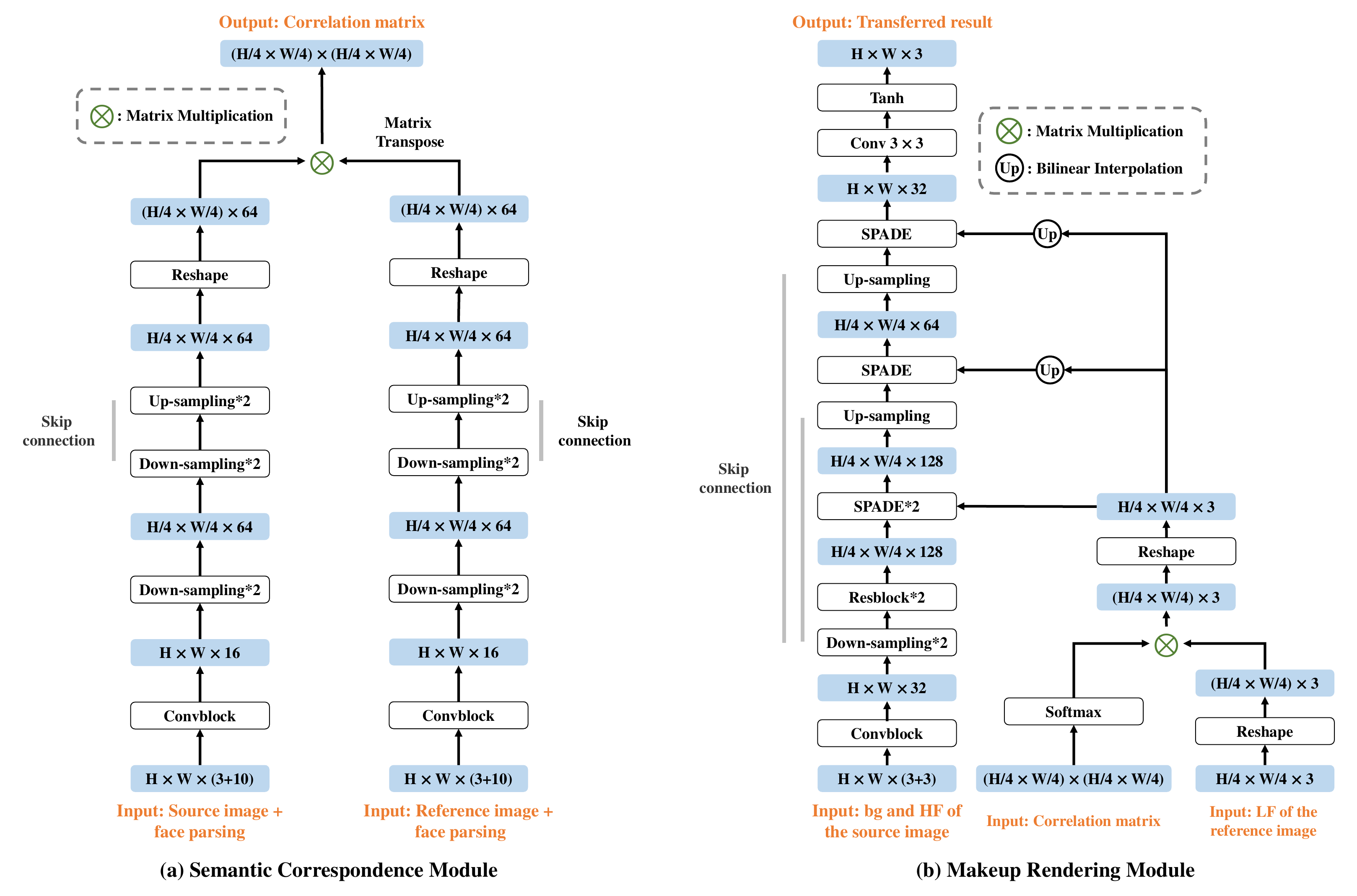}
	\caption{
		The architecture of the semantic correspondence module and makeup rendering module in CSD-MT. \emph{*n} indicates a stack of \emph{n} blocks.
	}
	\label{img:supply_network_details}
\end{figure*}

We design 5 basic network blocks to construct the generator $\mathcal{G}$ in our CSD-MT model, including Convblock, Down-sampling block,  Up-sampling block, Resblock, and SPADE, whose structures are shown in Figure \ref{img:supply_basic_blocks}. Based on these blocks, Figure \ref{img:supply_network_details} illustrates the architectures of the semantic correspondence module and the makeup rendering module, where the shape of each intermediate feature map is also presented. Note that, before feeding the input images into the semantic correspondence module, we concatenate them with their corresponding face parsing \cite{BiSeNet} maps to enhance the local semantic information of different facial parts (in our implementation, the face parsing maps of 10 semantic categories are utilized). To align with the target distribution, the proposed CSD-MT model adopts the same multi-scale discriminator $\mathcal{D}$ as in \cite{Pix2PixHD}, which consists of 3 scale-specific discriminators trained at 3 different image scales with an identical architecture.

\section{Training Details}
During the model training, each input image is manually resized to 256 $\times$ 256 pixels. In our color contrastive loss, four negative samples are generated for each transferred image, i.e., $N=4$ in Eq. (8). And the feature maps from the \texttt{relu\_1\_2}, \texttt{relu\_2\_2} layers of the pre-trained VGG19 model are used for calculating gram matrices (see Eq. (9)). For the hyper-parameters, we set $\tau=100$ in Eq. (2), $\alpha=0.1$ in  Eq. (4), and $\lambda_{trans}=1$, $\lambda_{cycle}=10$, $\lambda_{adv}=1$, $\lambda_{aug}=10$, $\lambda_{cts}=1$ in Eq. (10). We use the Adam \cite{Adam} optimizer with $\beta_{1}=0.5$ and $\beta_{2}=0.999$ for model training, the maximum number of training iterations is 500,000, the learning rate is 0.0002, and the batch size is 1.

\section{Parameter Size and Inference Speed}
In addition to the makeup transfer performance, we also compare the parameter size and inference time of CSD-MT with those of the competing methods. For a fair comparison, all the experiments are conducted on a single NVIDIA GTX 1660Ti GPU with 6GB RAM. From the results in Table \ref{table:Speed}, it can be seen that our CSD-MT model has the least number of parameters (6.94 M) and achieves the fastest inference speed (only 0.017 seconds for processing a pair of input images with a resolution of 256 $\times$ 256 pixels), which surpasses other benchmark methods by a large margin. This indicates the efficiency of our CSD-MT method.

\begin{table}[b]\small
	\begin{center}
		\begin{tabular}{l|c|c}
			\hline
			Methods & Parameters (M) & Inference Time (s)   \\
			\hline
			BeautyGAN \cite{BeautyGAN} & 9.42  & 0.039    \\
			PSGAN \cite{PSGAN} & 12.62& 0.218    \\
			SCGAN \cite{SCGAN}  & 15.30& 0.321    \\
			SpMT \cite{SpMT} & 333.67& 0.834    \\
			LADN \cite{LADN} & 27.00& 0.032    \\
			SSAT \cite{SSAT} & 10.48& 0.110     \\
			EleGANt \cite{EleGANt} & 10.27& 0.148     \\
			CSD-MT (ours)  & \textbf{6.94}  & \textbf{0.017}  \\
			\hline
		\end{tabular}
	\end{center}
	\caption{Comparisons of the parameter size and inference speed of CSD-MT and other methods. The number of parameters (M) and inference time (seconds) are calculated for different models when processing a pair of input images with a size of 256 $\times$ 256 pixels.}
	\label{table:Speed}
\end{table}

\begin{table}[h]\small
	\begin{center}
		\begin{tabular}{l|c|c}
			\hline
			\multirow{2}{*}{Parameter} & \multicolumn{2}{c}{Self-Aug PSNR/SSIM}  \\
			\cline{2-3}
			& Crop & Rotate \\
			\hline
			$\alpha=0.0$   & 23.77/0.830 & 22.12/0.791 \\
			$\alpha=0.1$   & \textbf{27.28/0.920} & \textbf{26.68/0.915}  \\
			$\alpha=0.5$   & 24.91/0.908 & 24.62/0.906 \\
			\hline
		\end{tabular}
	\end{center}
	\caption{Quantitative comparison of CSD-MT models trained with different $\alpha$ on the MT dataset.}
	\label{table:structure}
\end{table}

\begin{figure}[b]
	\centering
	\includegraphics[width=1.0\linewidth]{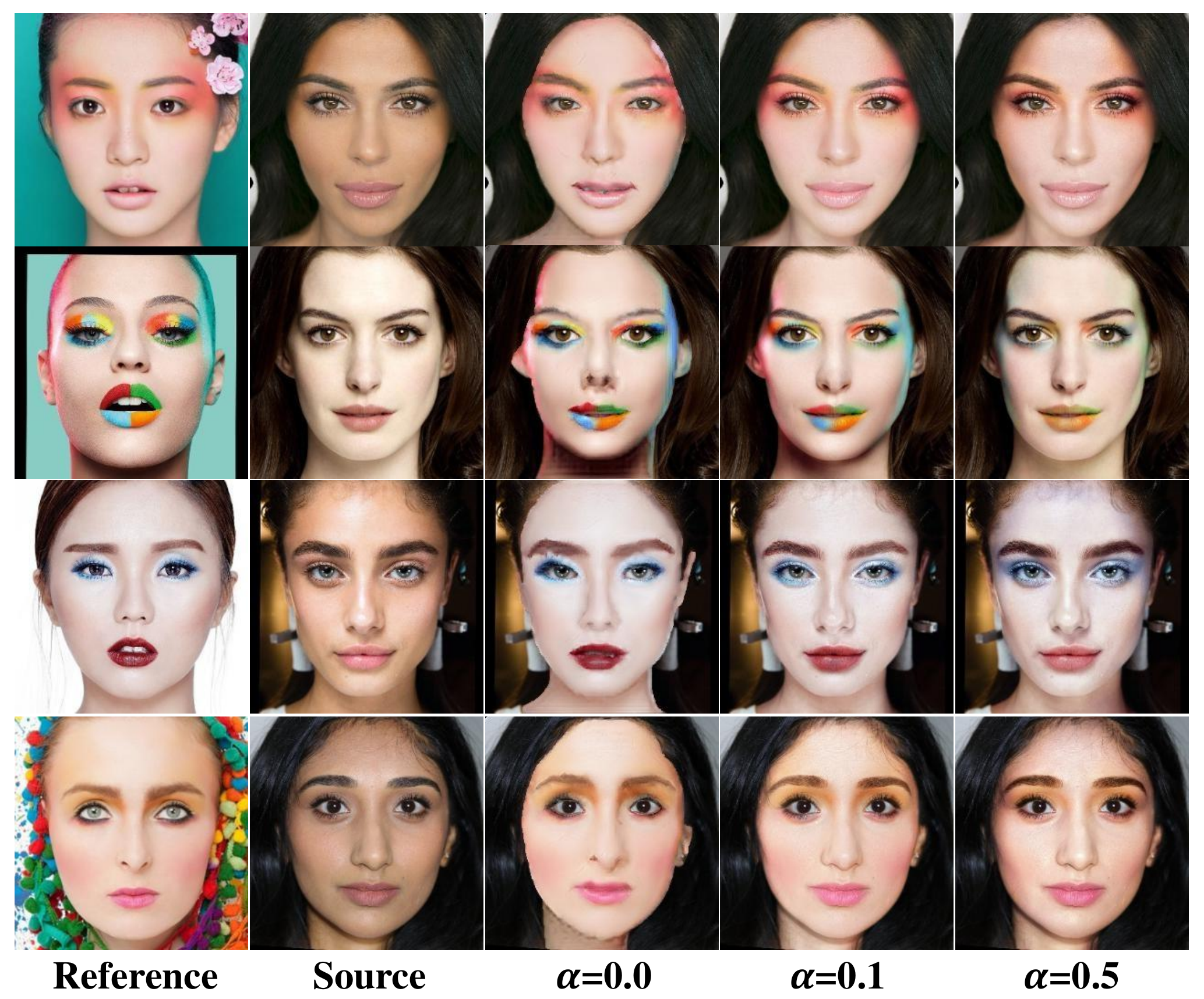}
	\caption{
		Qualitative comparison of CSD-MT models trained with different $\alpha$. $\alpha=0.1$ leads to the best transferred results.
	}
	\label{img:supply_trade-off}
\end{figure}

\section{Trade-off between Content and Makeup}
By minimizing the transfer loss $L_{trans}$ (see Eq. (4) in the main text), CSD-MT simultaneously preserves the content details in the source image ($L_{cont}$) and transfers the makeup information of the reference face ($L_{makeup}$). There is a trade-off between these two objectives, which is balanced by the importance parameter $\alpha$. To investigate the effect of this parameter, we compare the performance of CSD-MT models trained with different values of $\alpha$ (varying in $\{0.0, 0.1, 0.5\}$). Both quantitative and qualitative comparisons are conducted. As shown in Table \ref{table:structure}, the proposed CSD-MT method achieves the best self-augmented PSNR/SSIM results when $\alpha=0.1$ (27.28/0.920 and 26.68/0.915 on "Crop" and "Rotate" scenarios, respectively). Such phenomenon can also be found in Figure \ref{img:supply_trade-off}. When $\alpha=0.0$, the content objective $L_{cont}$ is removed from $L_{trans}$, so the trained model fails to retain the content information in the source images and generates unrealistic results. When the value of $\alpha$ increases to $0.5$, $L_{cont}$ dominates the transfer loss $L_{trans}$ and reduces the relative importance of $L_{makeup}$. As a result, the makeup styles of the reference faces, especially lipstick and powder blush, cannot be faithfully transferred.

\section{Comparison with Diffusion Models}
\begin{figure}[t]
	\centering
	\includegraphics[width=1\linewidth]{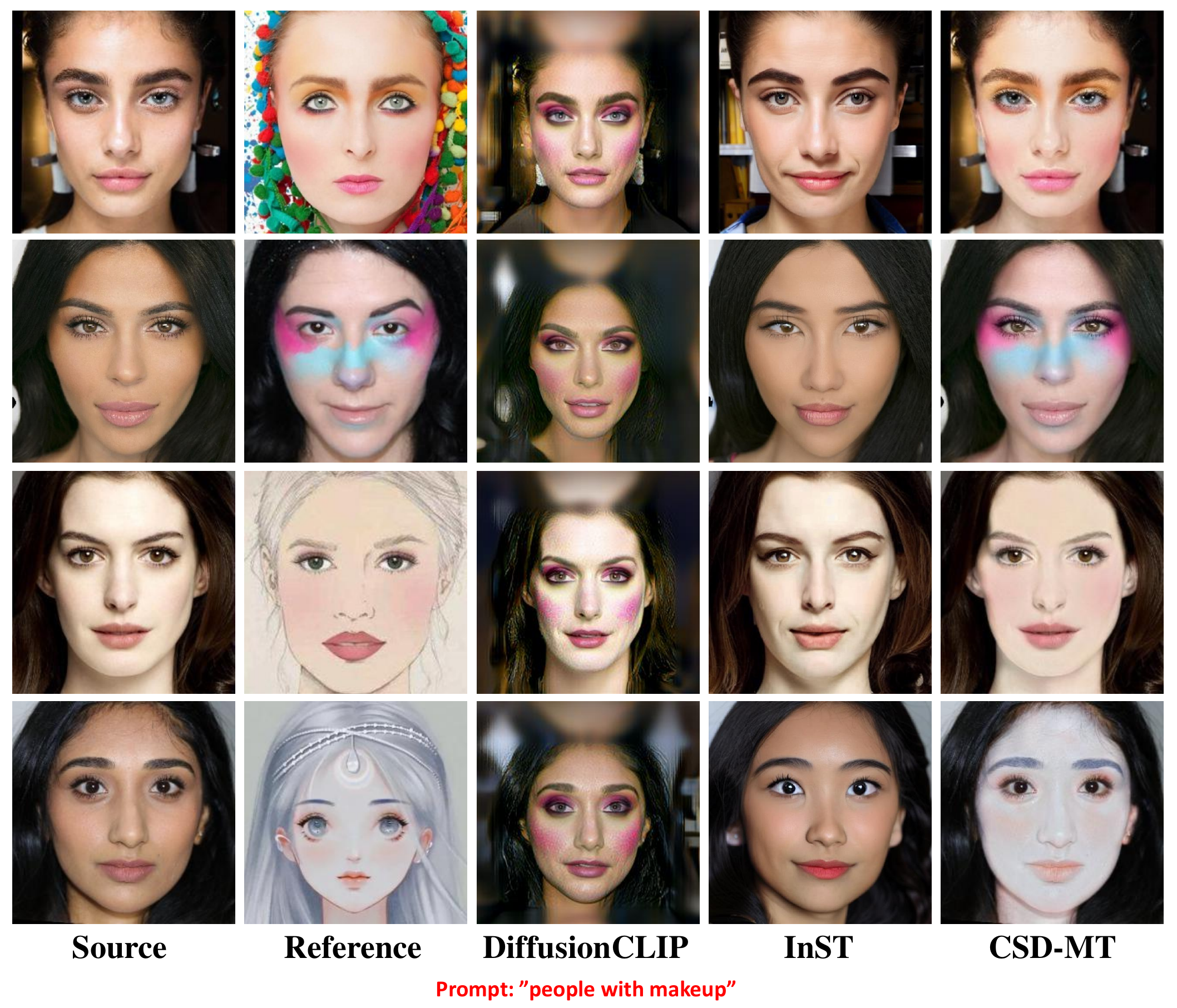}
	\caption{
		Qualitative comparison with diffusion models, including DiffusionCLIP \cite{DiffusionCLIP} and InST \cite{InST}.
	}
	\label{img:supply_comparison_with_diffusion}
\end{figure}

Recently, powerful diffusion models have been widely studied and become mainstream approaches for solving various image generation tasks. Therefore, we would also like to compare our CSD-MT method with diffusion models. Considering that there is currently no diffusion model specifically designed for the makeup transfer task, a text-guided generative diffusion model DiffusionCLIP \cite{DiffusionCLIP} and a style transfer diffusion model InST \cite{InST} are chosen as the benchmark methods. For DiffusionCLIP, since it is difficult accurately describe a specific makeup style in text, we use the prompt "people with makeup" as in \cite{DiffusionCLIP} to produce the final transferred results. From Figure \ref{img:supply_comparison_with_diffusion}, it can be seen that DiffusionCLIP usually introduces incorrect makeup information in the final outputs, since its generation process is mainly based on the text prompt instead of the reference image. As a style transfer method, InST not only fails to distill makeup styles from the reference image but also alters the content details of the source image. CSD-MT outperforms these two diffusion model based methods, again demonstrating its effectiveness and superiority.

\section{Makeup Control}
\subsection{Makeup Removal}
\begin{figure}[t]
	\centering
	\includegraphics[width=1.0\linewidth]{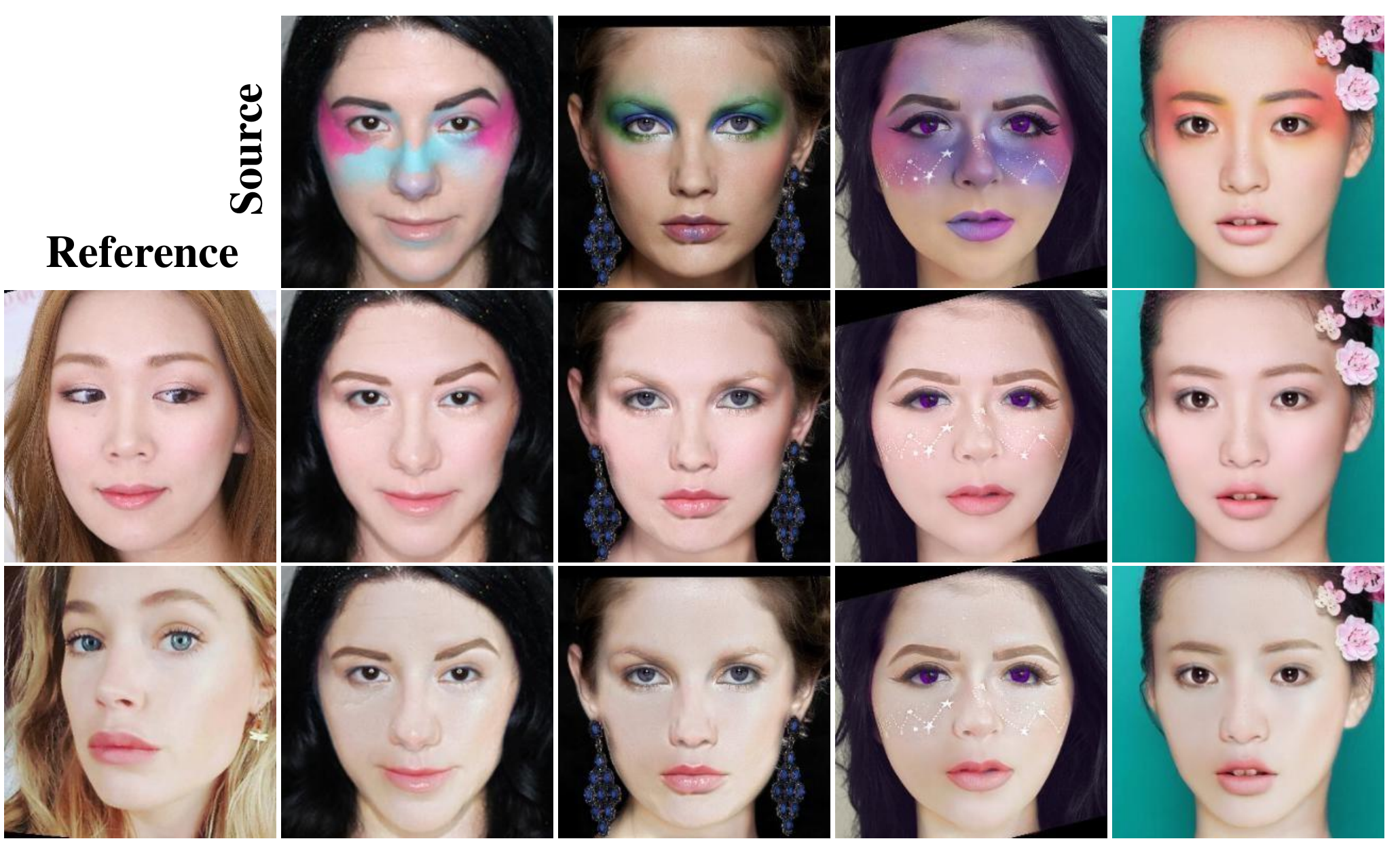}
	\caption{
		The makeup removal results generated by CSD-MT.
	}
	\label{img:supply_makeup_removal}
\end{figure}

Similar to \cite{DMT,LADN,SSAT}, by taking makeup images as the source inputs and non-makeup faces as the reference images, CSD-MT can also generate multiple makeup removal results, as displayed in Figure \ref{img:supply_makeup_removal}.

\begin{figure*}[t]
\centering
\includegraphics[width=0.98\linewidth]{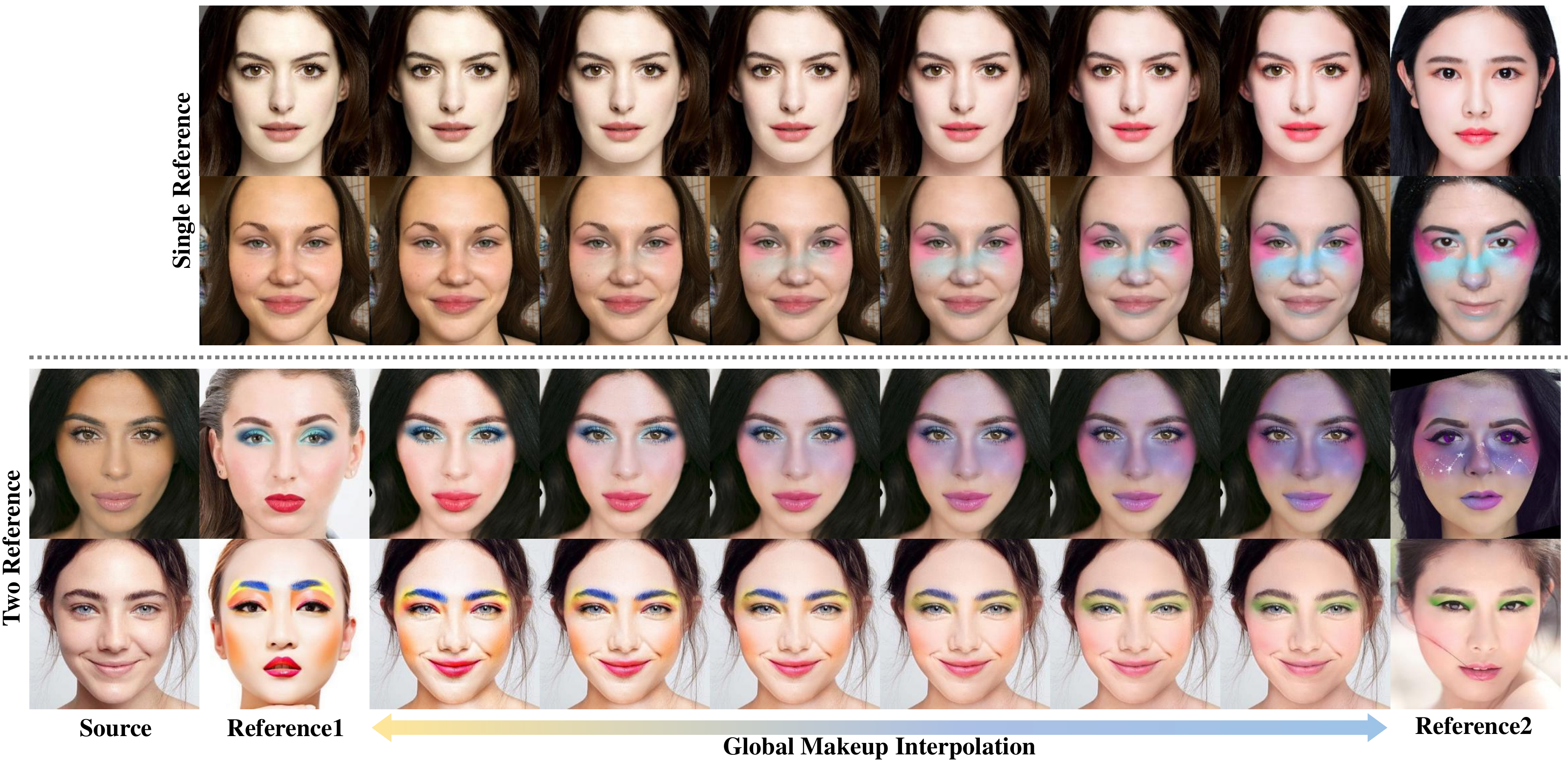}
\caption{
The illustration of global makeup interpolation. The first two rows are the result of a single reference image, the last two rows are the result of
two reference images.
}
\label{img:supply_global_makeup_interpolation}
\end{figure*}

\begin{figure*}[t]
	\centering
	\includegraphics[width=0.98\linewidth]{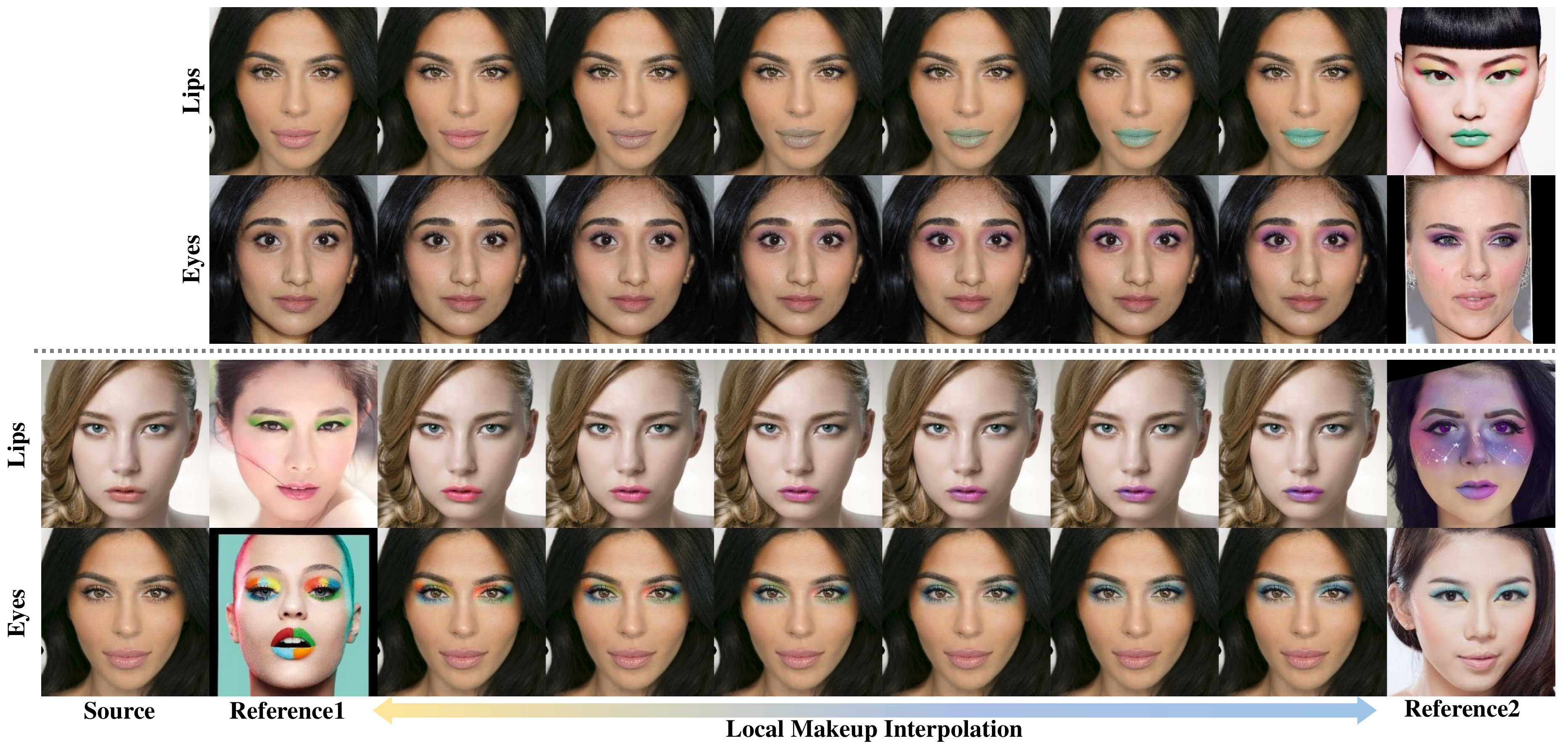}
	\caption{
		The illustration of local makeup interpolation. The odd rows are lipstick control, the even rows are eye shadow control.
	}
	\label{img:supply_local_makeup_interpolation}
\end{figure*}

\subsection{Global Makeup Interpolation}
In our proposed CSD-MT method, the makeup information are decoupled from the input images through frequency decomposition. This allows us to interpolate the makeup styles between two different reference faces by linearly
fusing their low-frequency (LF) components, as follows:
\begin{equation}\label{equ12}
	\begin{aligned}
		\hat{y}_{l}^{g\_inter}=& (1-\beta) \hat{y}_{l}^{1} + \beta \hat{y}_{l}^{2},\\
		\hat{x}^{g\_inter}=&G_{mr}([x_{bg},x_{h}],\hat{y}_{l}^{g\_inter}).
	\end{aligned}
\end{equation}
Here $\hat{y}_{l}^{1}$ and $\hat{y}_{l}^{2}$ are deformed LF components of two different reference images, respectively. By adjusting the value of $\beta$ from $0$ to $1$, CSD-MT can generate a series of transferred results. Their makeup styles will gradually change from that of one reference image ${y}^{1}$ to that of the other ${y}^{2}$. Moreover, by assigning the source image as ${y}^{1}$, we can control the degree of makeup transfer for a single reference input ${y}^{2}$. The global makeup interpolation results are shown in Figure \ref{img:supply_global_makeup_interpolation}.

\subsection{Local Makeup Interpolation}
In CSD-MT, the LF component of the reference image is deformed through the correlation matrix $M$, so that it can be semantically aligned with the source image. Such spatial alignment enables CSD-MT to implement the makeup interpolation within different local facial areas, which can be formulated as follows:
\begin{equation}\label{equ13}
\begin{aligned}
\hat{y}_{l}^{l\_inter}=& ((1-\beta) \hat{y}_{l}^{1} + \beta \hat{y}_{l}^{2}) \otimes Mask_{x}^{area} \\
+  &\hat{x}_{l} \otimes (1-Mask_{x}^{area}),\\
\hat{x}^{l\_inter}=&G_{mr}([x_{bg},x_{h}],\hat{y}_{l}^{l\_inter}).
\end{aligned}
\end{equation}
where $\otimes$ denotes the Hadamard product. $Mask_{x}^{area}$ is a binary mask of the source image $x$, indicating the local areas to be makeup, which can be obtained by face parsing. Figure \ref{img:supply_local_makeup_interpolation} visualizes the local makeup interpolation results within the areas around the lips and eyes, respectively, i.e., $area \in \{lip, eye\}$ for $Mask_{x}^{area}$. Similarly, we can also control the local makeup transfer degree of a single reference image by replacing the other reference input with the source image, as shown in the first two rows of Figure \ref{img:supply_local_makeup_interpolation}.

\begin{figure}[t]
	\centering
	\includegraphics[width=1.0\linewidth]{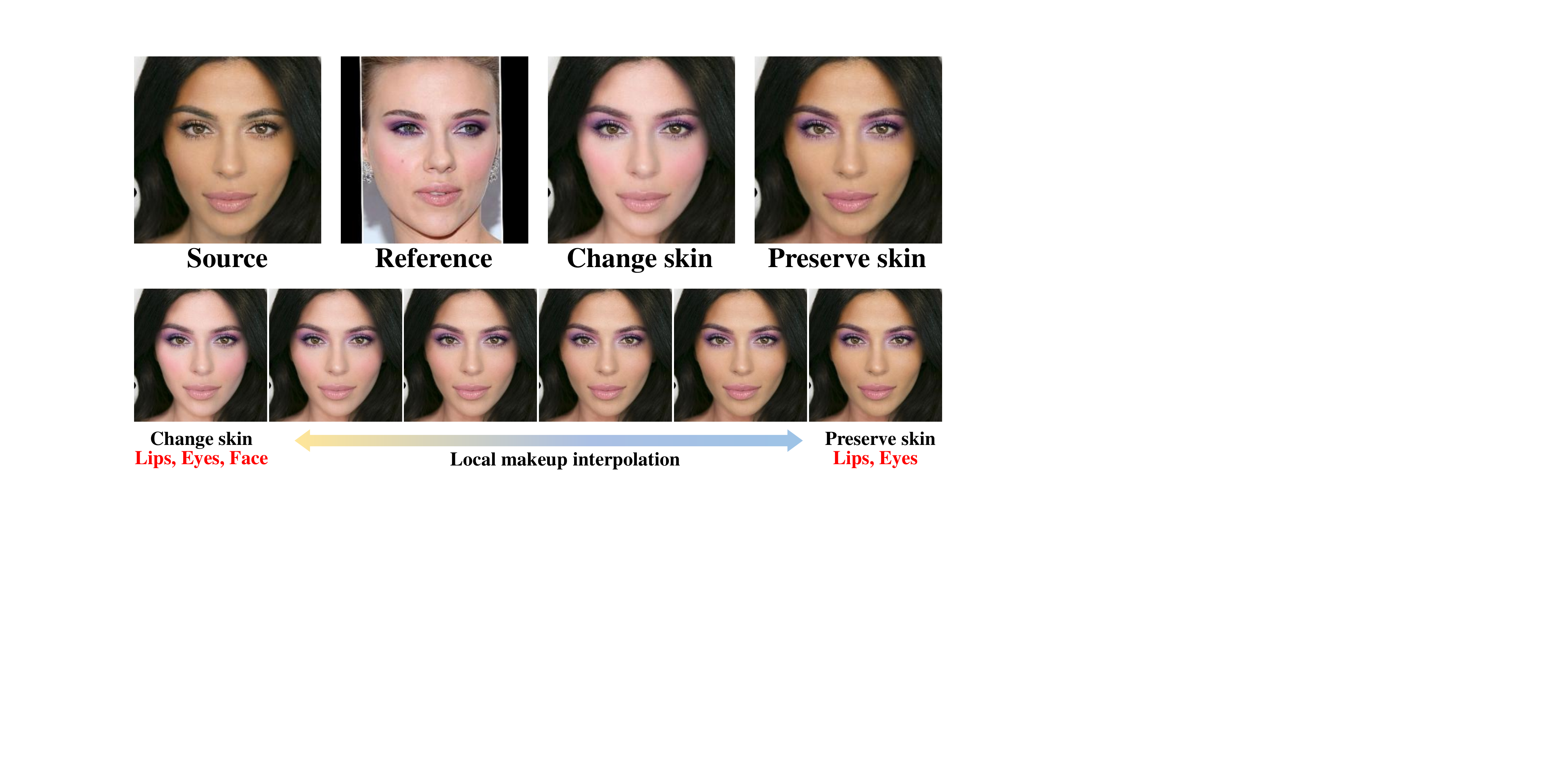}
	\caption{
		By default, our method CSD-MT transfers makeup to change the skin tone.
		Optionally, the local makeup transfer operation can preserve the original skin tone, and the local makeup interpolation can smoothly generate intermediate results.
	}
	\label{img:supply_skin_color}
\end{figure}

\noindent \textbf{Preserving Skin Tone.} Similar to previous approaches \cite{PairedCycleGAN,BeautyGAN,PSGAN,SCGAN,SpMT,LADN,SSAT,EleGANt}, CSD-MT assumes that the foundations and other cosmetics have already covered the original skin tone. Therefore, the skin color of the reference face is considered as a part of its makeup styles and is faithfully transferred to the final generated result, which may corrupt the content information in the source image. To alleviate this problem, we can perform the above-mentioned local makeup interpolation operation in the face region of the source image to preserve its skin tone. This procedure can be formulated as:
\begin{equation}\label{equ14}
\begin{aligned}
\hat{y}_{l}^{l\_skin}=& ((1-\beta) \hat{x}_{l} + \beta \hat{y}_{l}^{2}) \otimes Mask_{x}^{face} \\
+  &\hat{y}_{l}^{2} \otimes (1-Mask_{x}^{face}),\\
\hat{x}^{l\_skin}=&G_{mr}([x_{bg},x_{h}],\hat{y}_{l}^{l\_skin}).
\end{aligned}
\end{equation}
Here, $\hat{x}^{l\_skin}$ realizes the local makeup interpolation between the source image $x$ and the reference image ${y}^{2}$ within the face region in $x$, which is indicated by the mask $Mask_{x}^{face}$. The interpolation results are visualized in Figure \ref{img:supply_skin_color}. When $\beta=0$, $\hat{x}^{l\_skin}$ will not change the skin tone of $x$. And when $\beta=1$, Eq. (\ref{equ14}) degenerates to the standard makeup transfer process in CSD-MT, which will distill the makeup information (including the skin tone) from ${y}^{2}$ to $x$.

\subsection{Partial Makeup Transfer}
\begin{figure}[t]
	\centering
	\includegraphics[width=1.0\linewidth]{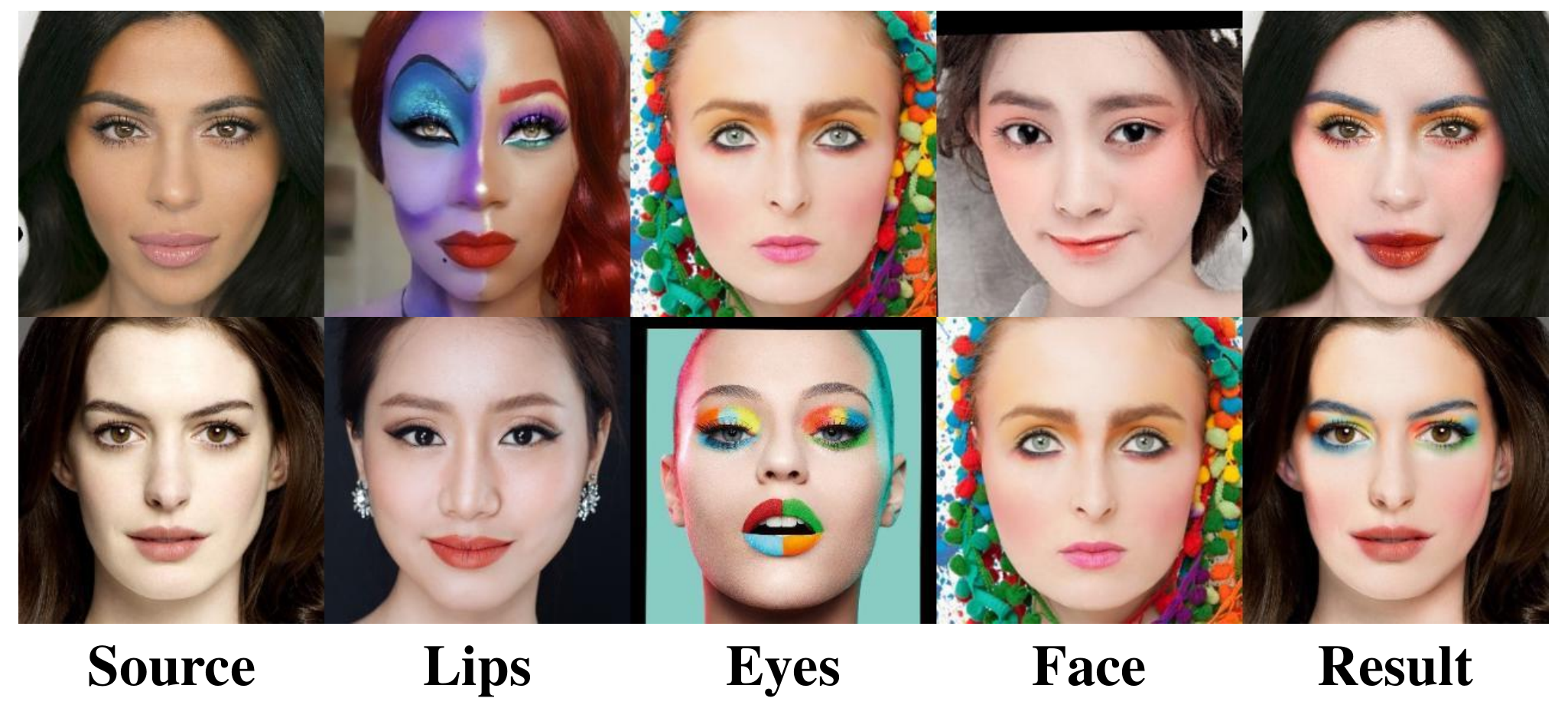}
	\caption{
		The results of partial makeup transfer. The results integrate the lips style from the second column, the eyes style from the third column, and the face style from the fourth column.
	}
	\label{img:supply_part_transfer}
\end{figure}

In addition, CSD-MT can integrate local makeup styles from different reference images for partial makeup transfer.
\begin{equation}\label{equ15}
\begin{aligned}
\hat{y}_{l}^{part}=& \hat{y}_{l}^{1} \otimes Mask_{x}^{lip} + \hat{y}_{l}^{2} \otimes Mask_{x}^{eye} \\
+ & \hat{y}_{l}^{3} \otimes Mask_{x}^{face}, \\
\hat{x}^{part}=&G_{mr}([x_{bg},x_{h}],\hat{y}_{l}^{part}).
\end{aligned}
\end{equation}
where $Mask_{x}^{lip}$, $Mask_{x}^{eye}$, $Mask_{x}^{face}$ are the lip, eye and face masks of the source image $x$.
The results of partial makeup transfer are shown in Figure \ref{img:supply_part_transfer}.

\subsection{Makeup Editing}
\begin{figure}[t]
\centering
\includegraphics[width=1.0\linewidth]{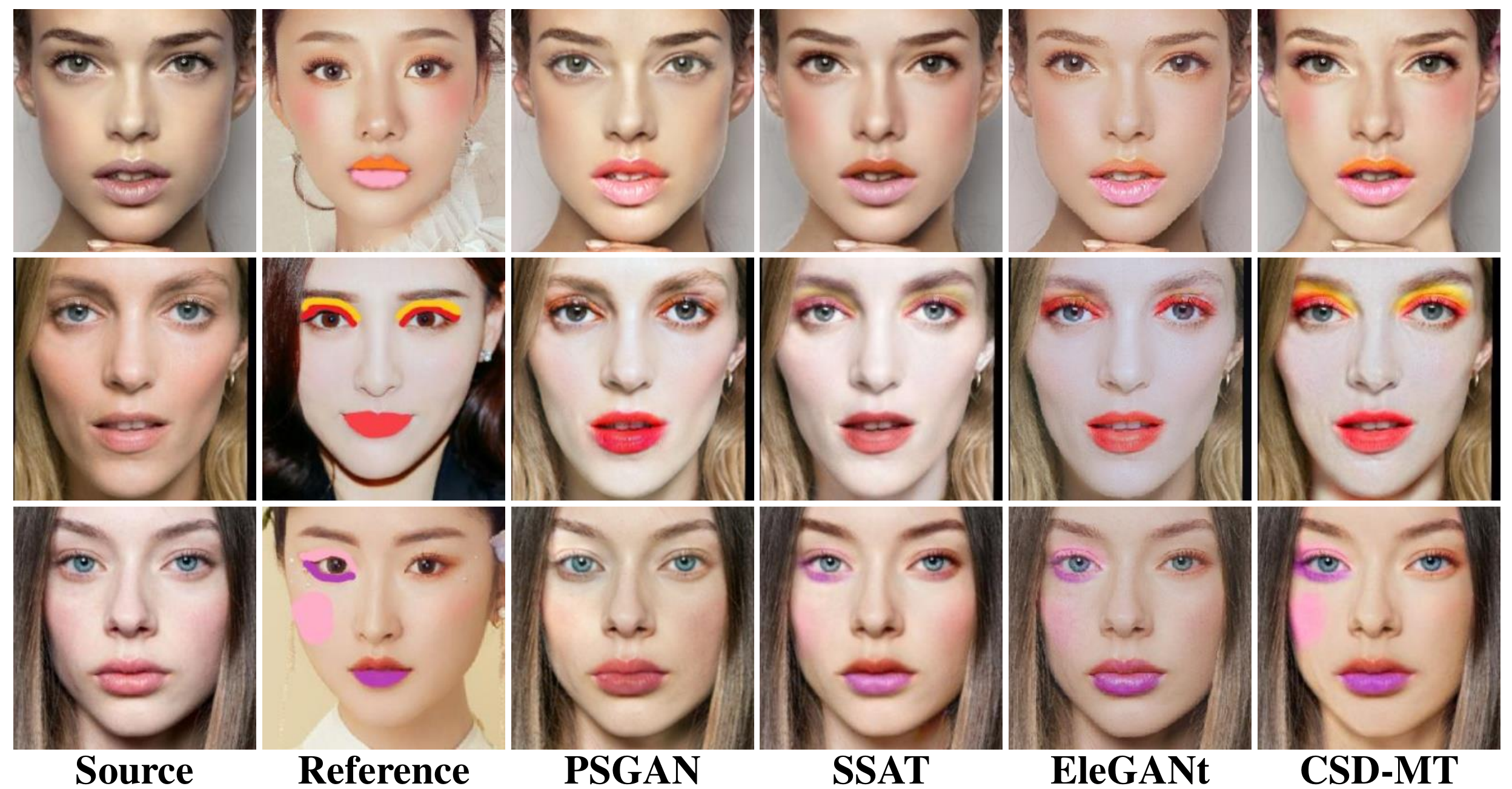}
\caption{
Comparison of makeup editing with different methods.
}
\label{img:supply_editing}
\end{figure}

CSD-MT also allows users to create their own customized makeup looks by editing the reference image. This editing process is simple and intuitive, the users only need to apply their preferred colors to any local area of the reference face. After that, our CSD-MT model is employed to transfer these user-edited makeup styles to the source images. As shown in Figure \ref{img:supply_editing}, CSD-MT generates better transferred results compared to other state-of-the-art methods.

\section{More Results}
\begin{figure*}[h]
\centering
\includegraphics[width=1\linewidth]{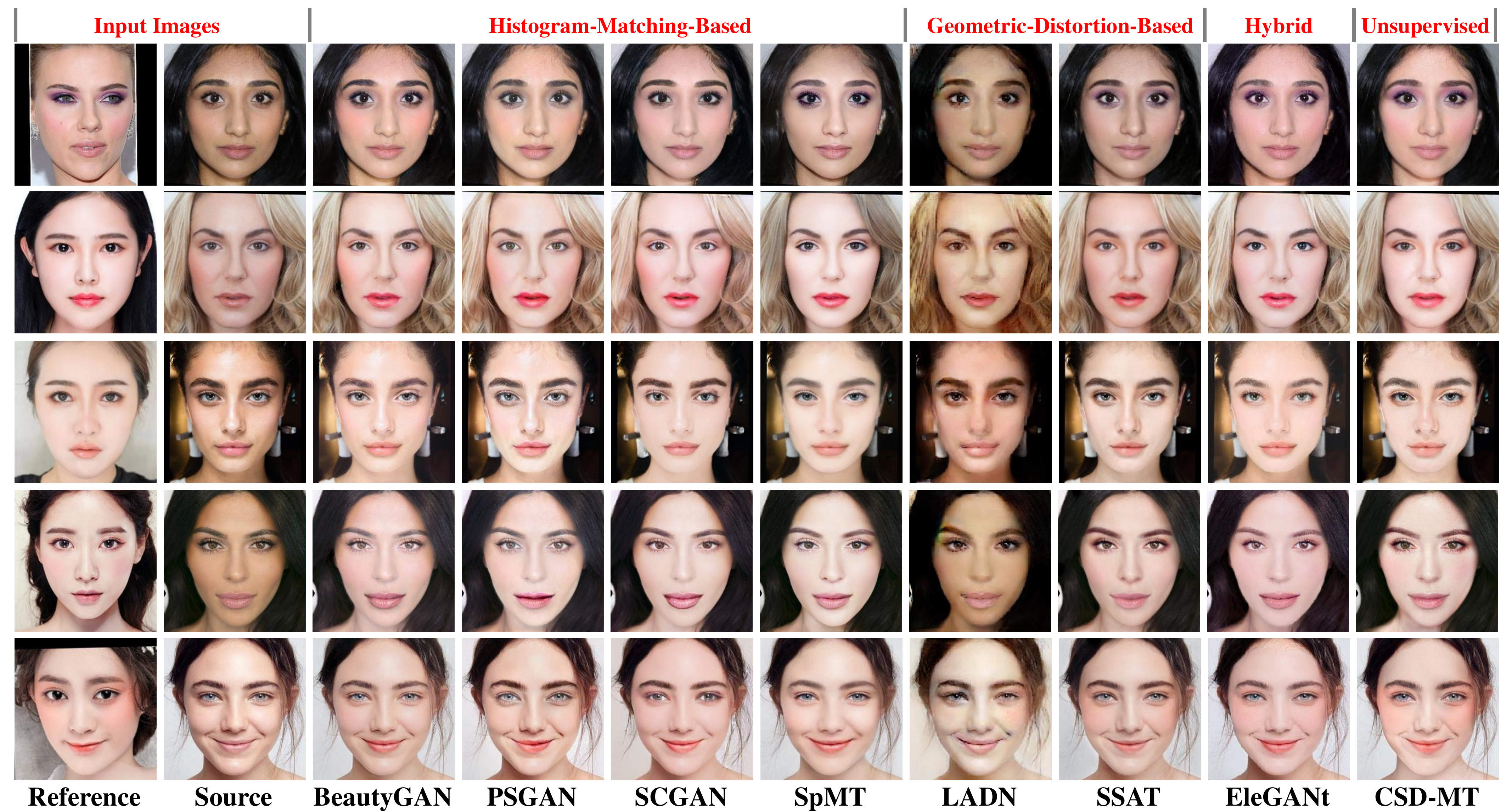}
\caption{
More qualitative comparisons between CSD-MT and state-of-the-art methods  \textbf{under simple makeup styles}.
}
\label{img:supply_qualitative_comparison1}
\end{figure*}

\begin{figure*}[h]
\centering
\includegraphics[width=1\linewidth]{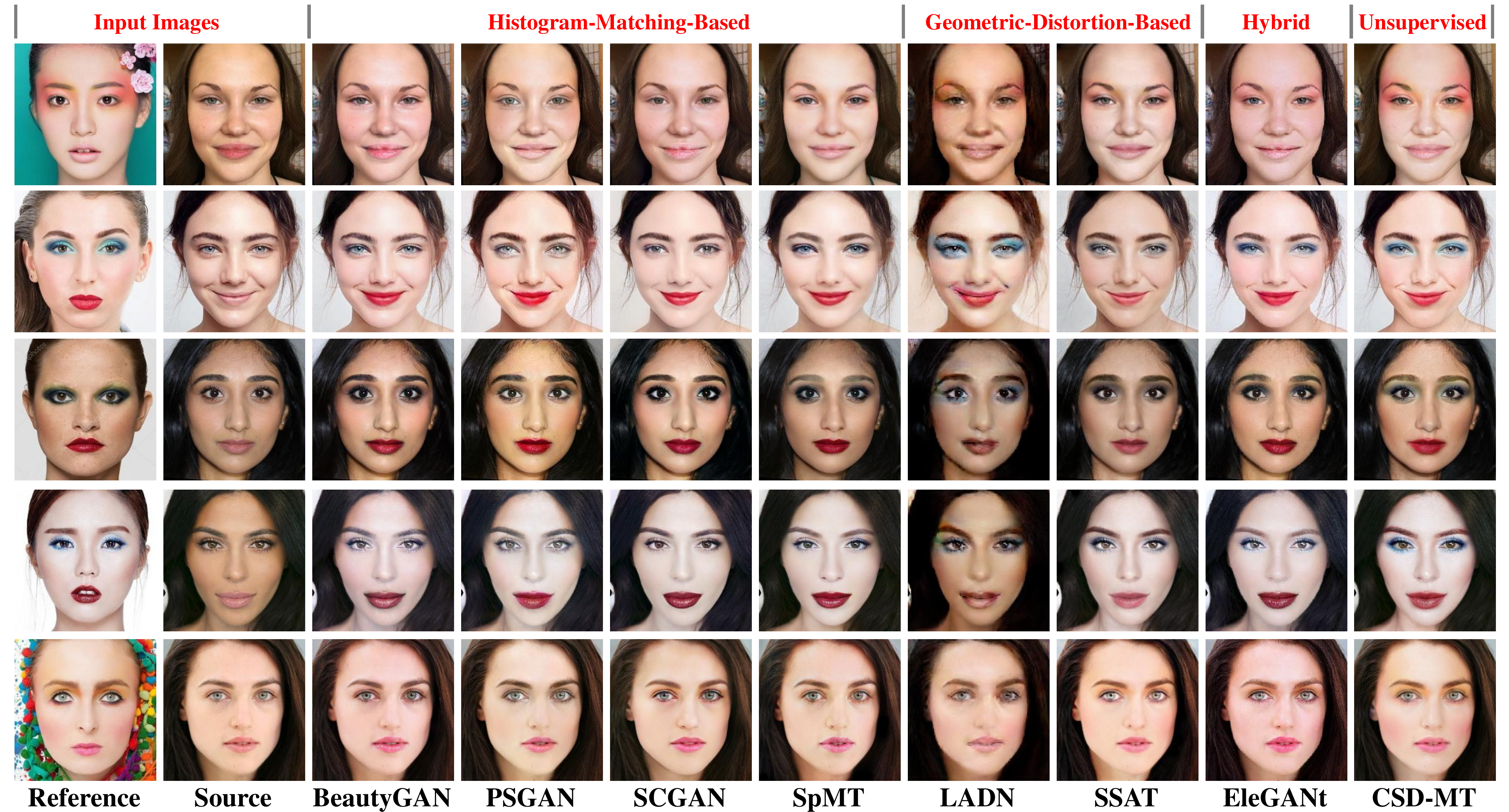}
\caption{
More qualitative comparisons between CSD-MT and state-of-the-art methods \textbf{under complex makeup styles}.
}
\label{img:supply_qualitative_comparison2}
\end{figure*}

\begin{figure*}[h]
\centering
\includegraphics[width=1\linewidth]{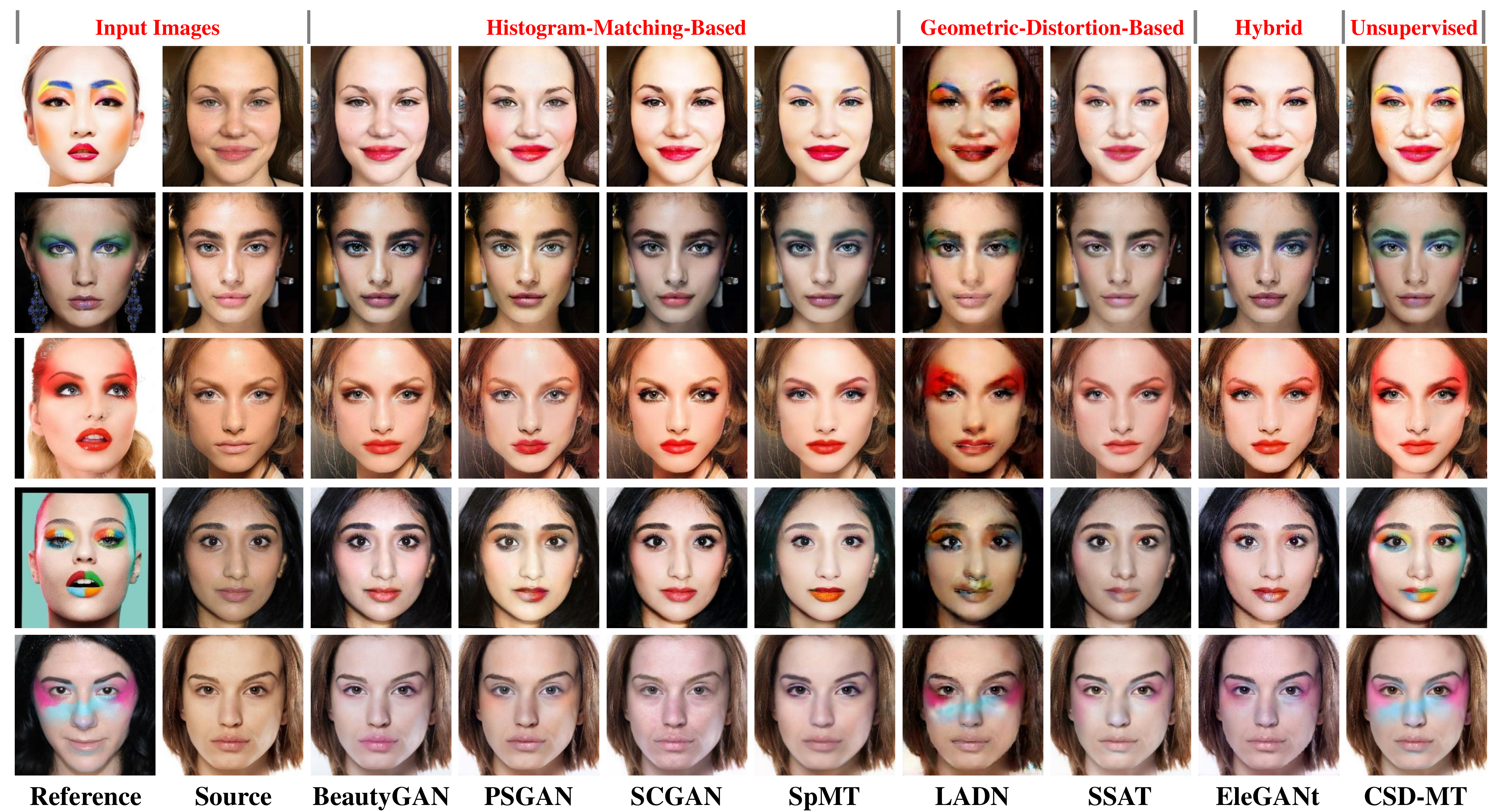}
\caption{
More qualitative comparisons between CSD-MT and state-of-the-art methods \textbf{under extreme makeup styles}.
}
\label{img:supply_qualitative_comparison3}
\end{figure*}

\begin{figure*}[h]
\centering
\includegraphics[width=1\linewidth]{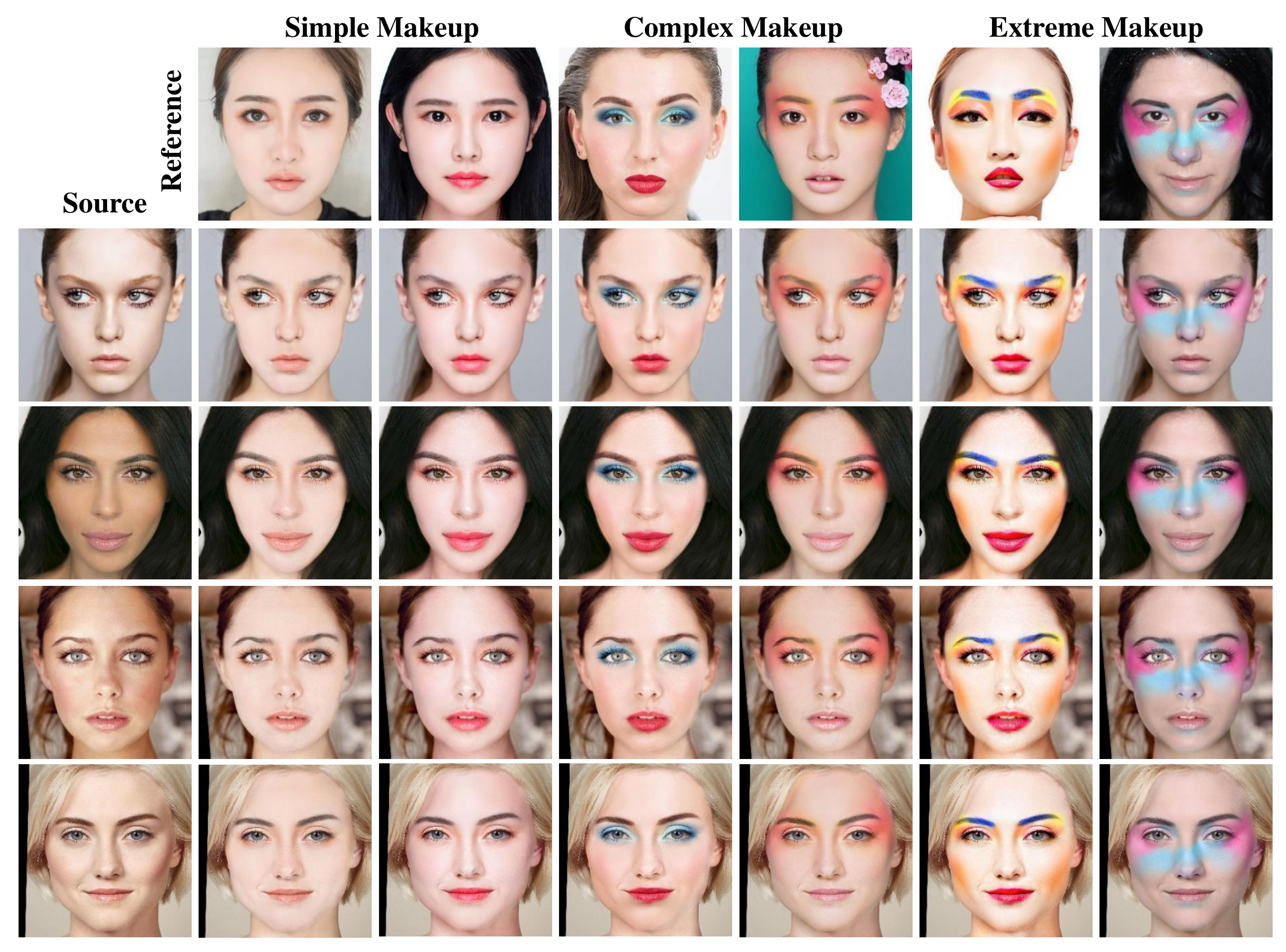}
\caption{
The makeup transfer results 1 of our CSD-MT under simple, complex, and extreme makeup styles.
}
\label{img:supply_our_results1}
\end{figure*}

\begin{figure*}[h]
\centering
\includegraphics[width=1\linewidth]{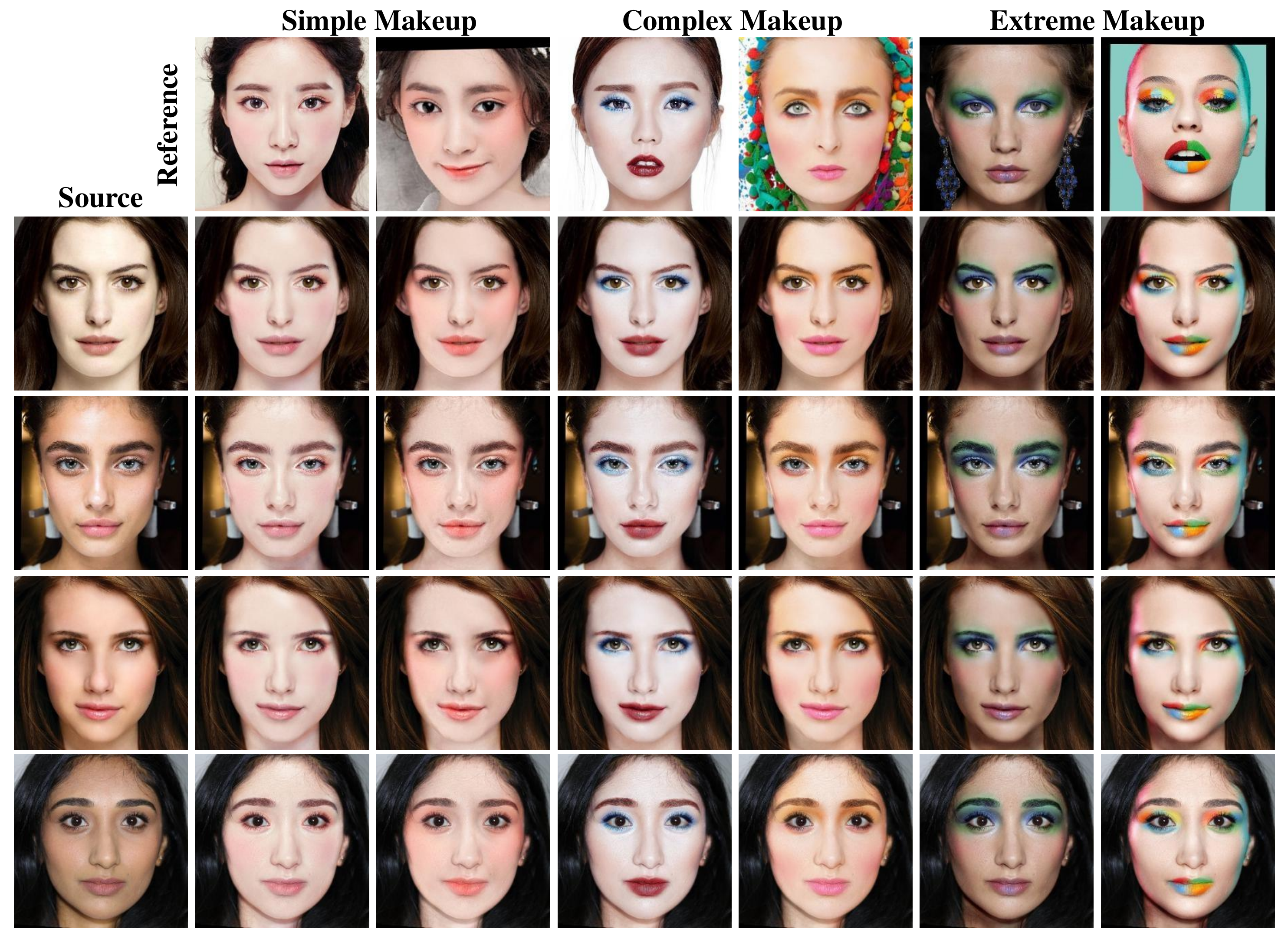}
\caption{
The makeup transfer results 2 of our CSD-MT under simple, complex, and extreme makeup styles.
}
\label{img:supply_our_results2}
\end{figure*}

\begin{figure*}[h]
\centering
\includegraphics[width=1\linewidth]{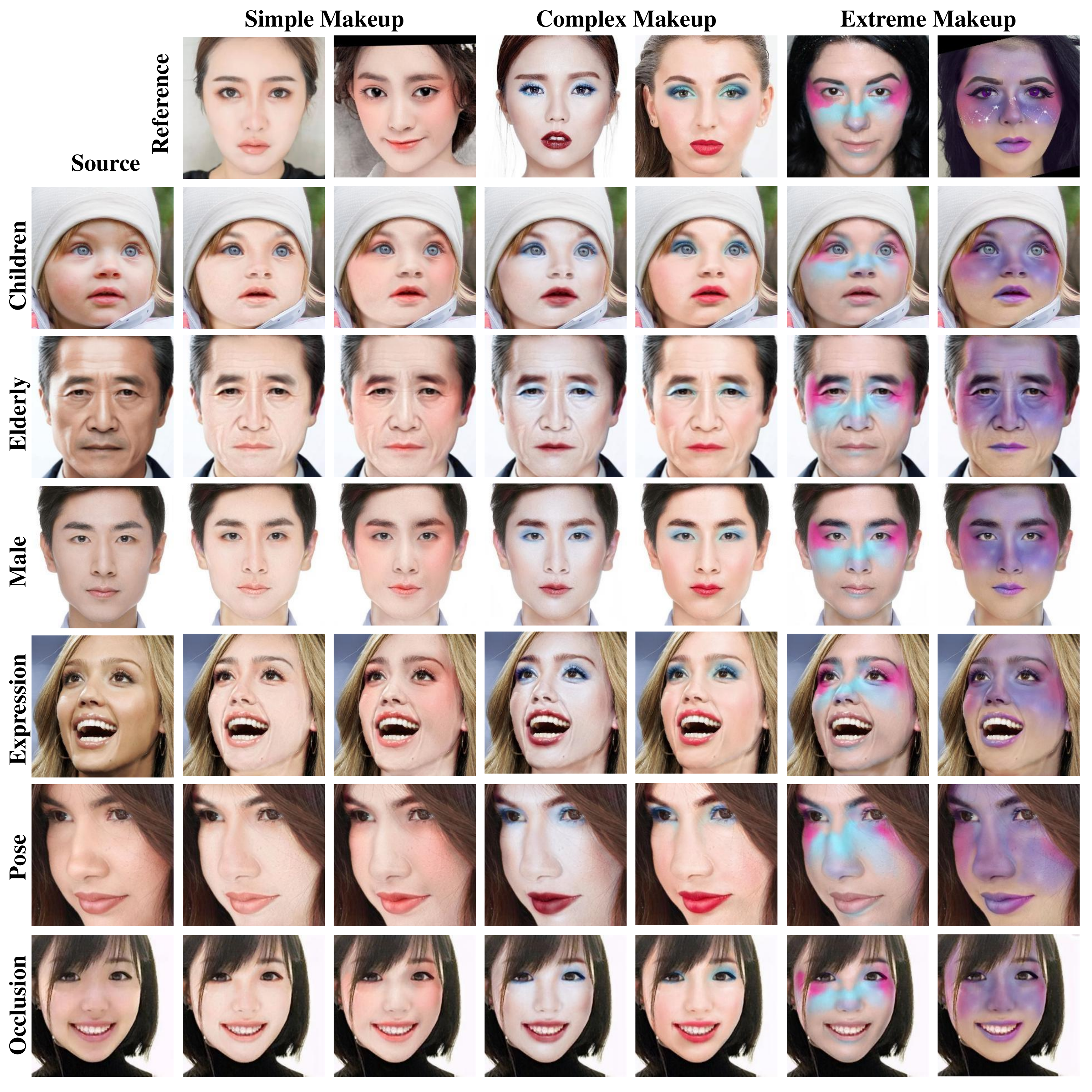}
\caption{
The robustness of CSD-MT in various complex scenarios.
}
\label{img:supply_robust}
\end{figure*}

\begin{figure*}[h]
\centering
\includegraphics[width=1\linewidth]{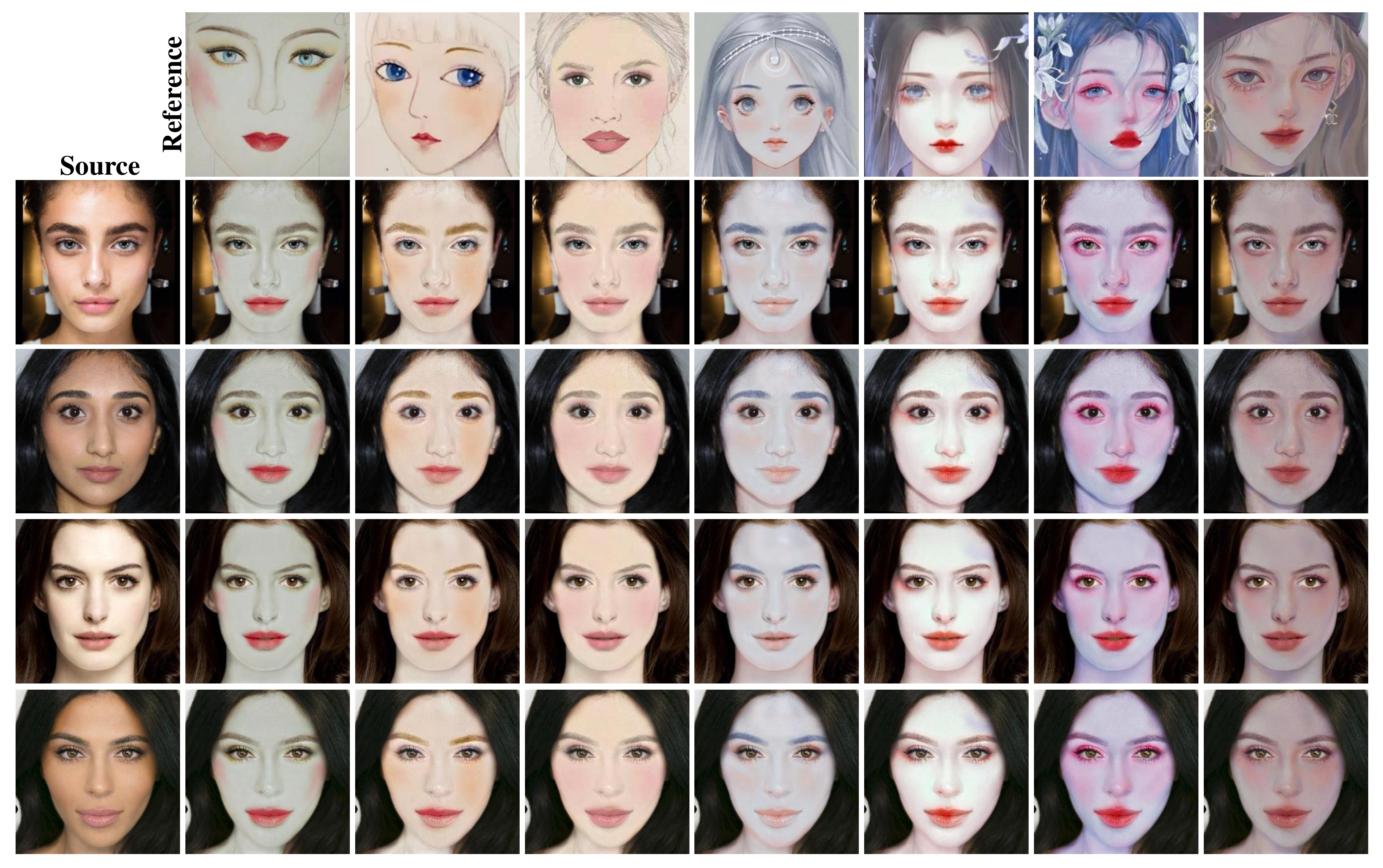}
\caption{
The generalization of CSD-MT in unsee anime makeup styles.
}
\label{img:supply_generalization}
\end{figure*}

\begin{figure*}[h]
\centering
\includegraphics[width=1\linewidth]{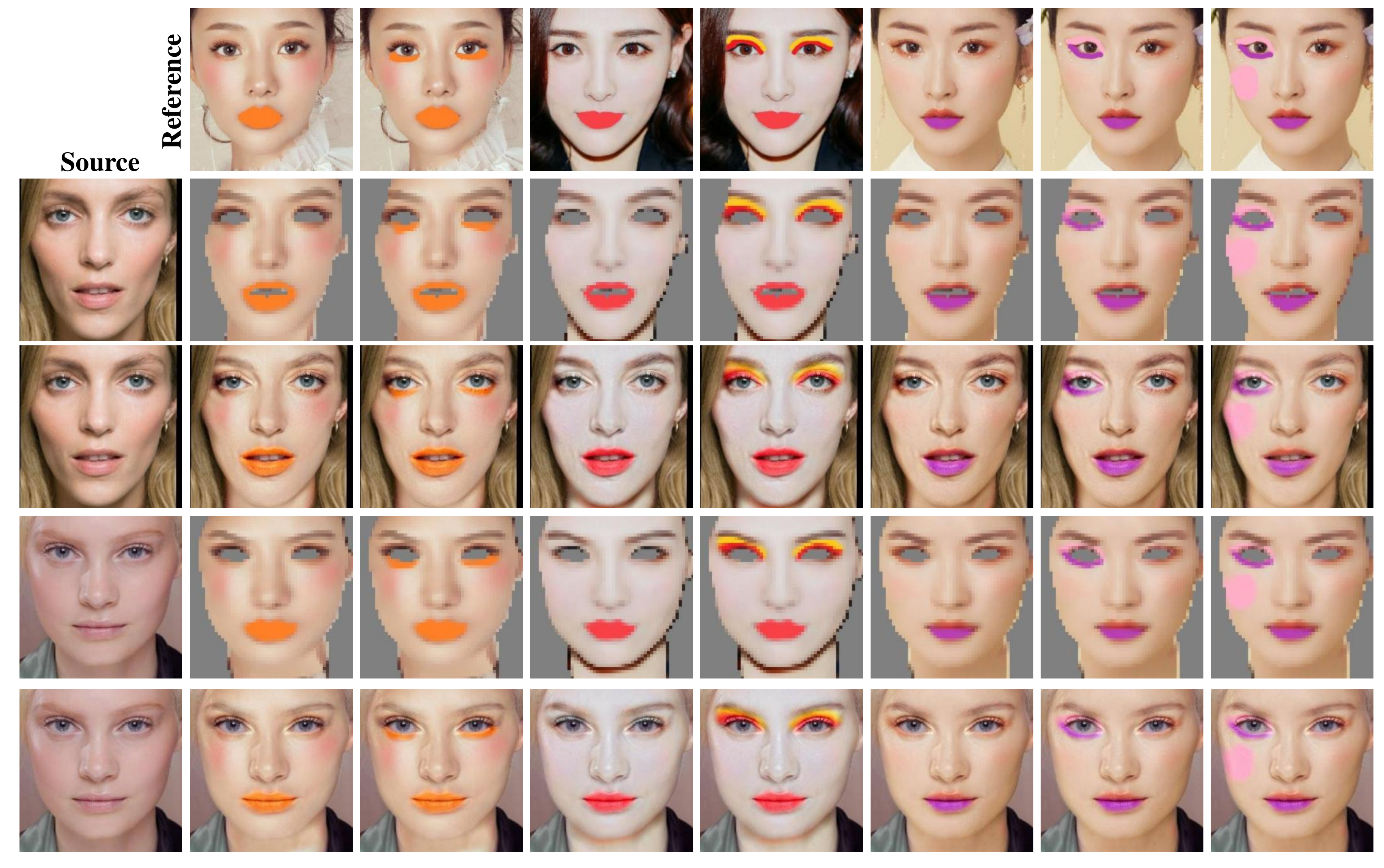}
\caption{
The controllability of CSD-MT in makeup editing. The deformed LF components are showcased to explain the makeup control mechanism of our approach.
}
\label{img:supply_makeup_editing}
\end{figure*}

Figure \ref{img:supply_qualitative_comparison1}, Figure \ref{img:supply_qualitative_comparison2}, and Figure \ref{img:supply_qualitative_comparison3} show more qualitative comparisons between CSD-MT and state-of-the-art methods under simple, complex, and extreme makeup styles, respectively.
More makeup transfer results of CSD-MT are shown in Figure \ref{img:supply_our_results1} and Figure \ref{img:supply_our_results2}.
Additionally, the robustness in various complex scenarios is demonstrated in Figure \ref{img:supply_robust}, the generalization ability to unseen makeup styles is shown in Figure \ref{img:supply_generalization}, and the control ability over makeup editing is illustrated in Figure \ref{img:supply_makeup_editing}.

\section{The Limitation}
\begin{figure*}[h]
\centering
\includegraphics[width=1\linewidth]{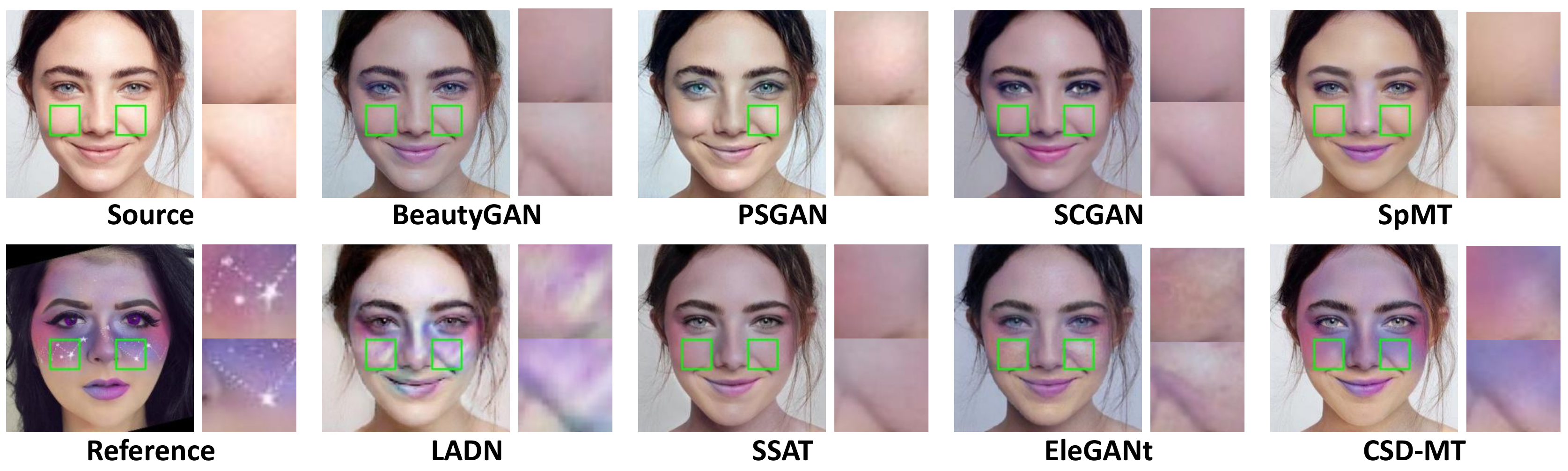}
\caption{
The limitation of our CSD-MT. We assume that the high-frequency (HF) component is more closely associated with the content details of face images.
As a result, our CSD-MT is ineffective in accurately reproducing the boundaries of some extreme makeup styles.
}
\label{img:supply_limitation}
\end{figure*}

In CSD-MT, we assume that the high-frequency (HF) component is more closely associated with the content details of face images.
With this assumption, CSD-MT preserves content details by maximizing the consistency of high-frequency information between the source image and the transferred result.
As a result, certain boundaries (HF information) of some extreme makeup are treated as content details rather than makeup style in CSD-MT.
Please refer to the makeup removal result in the fourth column of Figure \ref{img:supply_makeup_removal}.
At the same time, our CSD-MT is ineffective in accurately rendering the boundaries of some extreme makeup styles, as shown in Figure \ref{img:supply_limitation}.
In the future, our research will primarily focus on finding solutions to this problem.


